\documentclass[journal,twocolumn]{IEEEtran}
\usepackage{cite}
\usepackage{amsmath,amssymb,amsfonts}
\usepackage{algorithm}
\usepackage[noend]{algpseudocode}
\usepackage{tikz}
\usetikzlibrary{patterns,matrix,arrows}
\usepackage{graphicx}
\usepackage{textcomp}
\usepackage{xcolor}
\usepackage{url}
\usepackage{tikz}
\usepackage{pgfplots}
\usepgfplotslibrary{groupplots}
\usepackage{verbatim}
\usepackage{pgfplots}
\usepackage{subcaption}
\usepackage{graphicx}
\usepackage{hyperref}
\usepackage[inline]{enumitem}

\usepackage{etoolbox}\AtBeginEnvironment{algorithmic}{\footnotesize} 

\begin{document}
	\title{PassGoodPool: Joint Passengers and Goods Fleet Management with Reinforcement Learning aided Pricing, Matching, and Route Planning}
	\author{Kaushik Manchella$^{*}$, Marina Haliem$^{*}$, Vaneet Aggarwal, and Bharat Bhargava \thanks{* Equal Contribution. } \thanks{The authors are with Purdue University, West Lafayette, IN 47907 USA, email: \{kmanchel,mwadea,vaneet,bbshail\}@purdue.edu}\thanks{This work was presented, in part, at NeurIPS  Machine Learning for Autonomous Driving Workshop, Dec 2020.}}

	\maketitle
	
\begin{abstract}
The ubiquitous growth of mobility-on-demand services for passenger and goods delivery has brought various challenges and opportunities within the realm of transportation systems. As a result, intelligent transportation systems are being developed to maximize operational profitability, user convenience, and environmental sustainability.  The growth of last mile deliveries alongside ridesharing calls for an efficient and cohesive system that transports both passengers and goods. 
Existing methods address this using static routing methods considering neither the demands of requests nor the transfer of goods between vehicles during route planning. In this paper, we present a dynamic and demand aware fleet management framework for combined goods and passenger transportation that is capable of (1) Involving both passengers and drivers in the decision-making process by allowing drivers to negotiate to a mutually suitable price, and passengers to accept/reject, (2) Matching of goods to vehicles, and the multi-hop transfer of goods, (3) Dynamically generating optimal routes for each vehicle considering demand along their paths, based on the insertion cost which then determines the matching, (4) Dispatching idle vehicles to areas of anticipated high passenger and goods demand using Deep Reinforcement Learning (RL),  (5) Allowing for distributed inference at each vehicle while collectively optimizing fleet objectives. Our proposed model is deployable independently within each vehicle as this minimizes computational costs associated with the growth of distributed systems and democratizes decision-making to each individual.  Simulations on a variety of vehicle types, goods, and passenger utility functions show the effectiveness of our approach as compared to other methods that do not consider combined load transportation or dynamic multi-hop route planning. Our proposed method showed improvements over the next best baseline in various aspects including a 15\% increase in fleet utilization and a 20\% increase in average vehicle profits.

\end{abstract}


\section{Introduction}

\vspace{-0.1in}
\subsection{Motivation}
Ride Sharing market is expected to increase by a 20\% compound annual growth rate (CAGR) in the next few years \cite{policyadvise}. Further, about 90\% of retailers are expected to shift to crowdsourced delivery by 2028 \cite{zebra}. Platforms have three stakeholders: the offering company, the driver, as well as the customer. Customers seek convenience while drivers and the ride-hailing platforms seek profits \cite{hahn2017ridesharing}. Hence an ideal solution would be one that maximizes benefits for all these parties in providing passenger and goods transportation. Additionally, MoD services have proven potential to improve the efficiency and sustainabilty with vehicle sharing \cite{shaheen2018shared}. Urban traffic congestion, pollution, and energy expenditure are a few of the adverse side effects of transportation systems that stand to be improved by the adoption of intelligent transportation systems \cite{metz2018developing}. Route planning for mobility-on-demand (MoD) services has also proven to be an opportunity for optimization \cite{bei2018algorithms} towards more profitable transportation models. 
In the development of technology for vehicle networks, another challenge to deal with is the computational challenges that arise due to scaling the number of vehicles \cite{InternetofAutonomousVehicles,lei2020efficient}. To support quick inferences during driving, vehicles are increasingly being equipped with computational resources for in-house deep learning inference \cite{molanes2018deep}. Considering this, we present a distributed algorithm for fleet management where each vehicle is able to learn its own policies and make its own decisions while contributing to the optimization of fleet objectives.
The key aspects of the fleet management system include the passenger-vehicle and goods-vehicle assignment, efficient route for the vehicle to pick up and drop the passengers and goods, efficient pricing to the passengers, and the dispatching of idle vehicles. {All these problems are inter-connected and influence each other, and thus cannot be studied independently.  Combining passengers and goods in one framework can potentially optimize the fleet utilization; however, they impose significant challenges. }As a special case, the route-planning problem (Given a set of vehicles $V$ and passenger/goods requests $R$, designing a route that consists of a sequence of pick-up and drop-off locations for a given vehicle) is NP-hard \cite{haliem2020distributed}. Since multiple sub-problems are discrete optimization problems with large number of decision variables, a computationally efficient architecture and framework is essential for the joint pricing, matching, route-planning, and dispatching, which is the focus of this paper.  This joint framework brings novelty in different components as well as strong integration that can guide future research in terms of the overall system rather than performance of individual sub-systems.

\vspace{-0.1in}
\subsection{Related Work}
{\bf{Dynamic Ridesharing:}} Model-free approaches have been proven to be efficient in the space of urban mobility for vehicle dispatching. The authors of \cite{oda2018movi} first introduced a distributed approach to vehicle dispatching using Deep Reinforcement learning, where pooling of passengers was not considered. The authors of \cite{al2019deeppool} expanded that to consider pooling which leads to significantly improved fleet performance in terms of accept rate, customer waiting time and fleet utilization. {However, all of these
approaches did not explicitly model pricing strategies which in fact are key drivers of the MoD marketplace. Additionally, the aforementioned approaches use naive matching approaches and left significant room for improvement in route-planning for vehicles. } Efficient matching and pricing in addition to the dispatch were further considered in \cite{haliem2020distributed}. {While this showed promising results, it did not consider layover-based multi-hop approaches in route planning and the framework only addresses passenger deliveries.}

\textbf{Joint Pricing and Matching:}
Regarding the pricing problem, there are various ways to divide the trip costs among the rideshare partners \cite{AGATZ20111450},  \cite{WANG2019122},\cite{zhang2020pricing}. 
However, none of the works in literature combine pricing with matching, they study them as two separate problems. Grouping requests together greatly affects the pricing decisions and vice-versa. Ascertaining grouping together requests that are not going in opposite directions has not been addressed in literature.  Most works in literature assume that vehicles are told which passengers go together, by taking near-by pickup locations together (e.g., \cite{xu2020efficient,  al2019deeppool}) and thus are at the risk of grouping rides going in opposite directions. 
In this paper, we propose a model-free technique for the ride-sharing pricing problem. In contrast to the model-based approaches in literature \cite{3,4,5,6}, our proposed approach can adapt to dynamic distributions of customer and driver preferences.  At the same time, we incorporate the pricing aspect to our matching algorithm which would prioritize rides with significant path-intersections to be grouped together. This provides an efficient dynamic pricing-aware matching algorithm for the ridesharing environment by managing the prices for passengers based on the distance traveled due to serving this passenger (i.e., inserting this request in the vehicle's current route). The intuition is that the insertion cost of a request leading to an opposite direction to any of the previously inserted requests will be very high, and thus will be highly unlikely to be inserted into this specific route.  In this manner, the matching, route planning, and pricing components work in tandem with each other to reach a common ground solution that is favorable by and profitable to both parties.

{\bf Joint Passenger Goods Ridesharing:} Specific to the passenger-vehicle and goods-vehicle assignment in route planning, it has been shown in the literature that the ``insertion operation" is an effective method to solve NP-hard dynamic assignments \cite{bei2018algorithms,zheng2017online,DARP}. However, these approaches were only limited to two requests per vehicle. Further, these approaches don't consider goods transportation. Recently, the authors of \cite{haliem2020distributed} leveraged the insertion-operation and extended matching to consider vehicle capacity constraints. However, this approach can't be applied directly since in our case, the goods can reach the destination in multiple hops, and the pricing for goods is based only on distance not the insertion cost. {In \cite{LI201431}, the authors formulated the freight insertion problem (FIP); however, they don't consider multi-hop routing that allows for further optimization. }

The authors of \cite{manchella2020flexpool} considered joint dispatch policy for goods and passengers. However, the approach uses greedy matching policies and do not consider any pricing of the passengers.{This paper proposes efficient inter-connected decisions of pricing, matching, and route-planning that influences the rewards in the dispatch problem, which is solved using a reinforcement learning approach. Specifically, we propose novel approaches by considering a combined workload of passengers and goods along with an improvement in route-planning with multi-hop considerations. 
The goal of our approach is to influence the customer and vehicle utility functions, utilizing an optimal dispatching framework, to (i) achieve convenient pricing and matching decisions, (ii) optimize vehicles' route planning to deliver both passengers and goods using multi-hop routing, and (iii) re-position idle vehicles to areas of predicted high demand.} 

\vspace{-0.1in}
\subsection{Contributions}
The key contributions of the paper are as follows:
\begin{enumerate}[leftmargin=*]
\item This paper provides an integrated passenger and goods transportation system, that allows for pooling capabilities of the passengers, and multi-hop transfer for the goods. The proposed system integrates efficient pricing to the passengers, assignment of passengers and goods to the vehicles, dynamic route planning for vehicles, and dispatching of vehicles when they experience idle times. 

\item This paper provides a dynamic multi-hop allocation algorithm that decides on where to best drop off goods for layovers along their journey. Our approach aims to minimize the driving time of vehicles while minimizing travel time on delivering goods.

\item A distributed pricing approach is proposed for the combined passengers and goods transportation problem. In this algorithm, passengers and vehicles negotiate for the best suitable price based on the preferences and utility of both parties. Further, vehicles are offered a price for goods delivery orders based on the distance gained towards the final destination. 

\item The proposed approach integrates a demand-aware insertion based route planning algorithm for both goods and passengers that is able to scale up to the maximum capacity of each vehicle. This algorithm takes into account the near-future demand as well as the profit margin associated with each delivery to improve route-planning by preferring rides that are along the vehicles' current route plans.

\item This paper provides a distributed dispatch policy for joint passengers and goods delivery using reinforcement learning. This approach integrates the reward functions from the pricing, matching, and route planning components, and thus achieves inter-connected decision-making among them. The optimization problem for dispatch aims to maximize vehicle profit and request acceptance while minimizing customer waiting time, idle driving time of vehicles, total number of vehicles on the road (to reduce traffic and congestion), and fuel consumption.



 \end{enumerate}
Using real-world taxi-trip, and goods-order datasets, we simulate experiments with our joint approach that shows significant improvements in (i) request accept rate (ii) vehicle profits (iii) idle driving. Our proposed method showed improvements in various aspects including an approximate 10\% increase in accept rate, 20\% increase in average vehicle profits, and approximately 20\% decrease in average idle driving time of the fleet and improving fleet utilization by 15\%. We note that each of the components of our framework are crucial to optimizing the fleet objectives as a whole.  
To the best of our knowledge, this is the first work combining pricing, matching, dispatching and route planning for joint passenger and goods transportation.

\vspace{-0.1in}
\subsection{Organization}
The rest of the paper is organized as follows.  Section \ref{System} describes the system model. Section \ref{Proposed_Architecture} provides the architecture and high level algorithm. Section \ref{Routing} discusses the route planning methodology including the multi-hop and insertion cost algorithms. Section \ref{Pricing} discusses the pricing strategies. Section \ref{Dispatching} describes the Distributed DQN approach for dispatching idle vehicles. Section \ref{Experiments} evaluates and discusses the results obtained from the experiment. Section \ref{Conclusions} concludes this work with discussion on future directions. 

\subsection{Abbreviations and Acronyms}\label{AA}
MoD: Mobility-on-Demand. DARM: Demand Aware Routing and Matching. DPRS: Distributed Pricing Approach for Ride-Sharing.  RL: Reinforcement Learning. DQN: Deep Q-Network. Conv-Net: Convolutional Neural Network.

\section{System Model}\label{System}
In this section, we discuss the key components of our system along with the key objectives.  In addition, the model parameters and notations are also provided. 

\vspace{-0.1in}
\subsection{{Proposed Architecture}}
\begin{figure}
	\includegraphics[width=.48\textwidth]{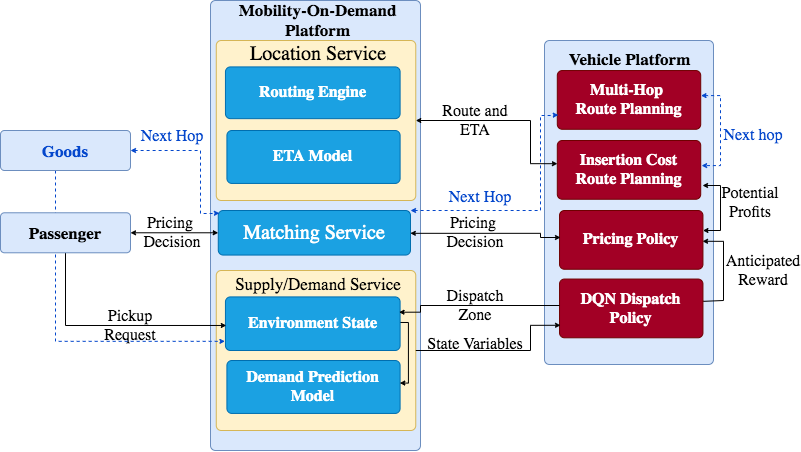}
	\caption{An illustration of the framework architecture. The image shows the various actors, services, and interactions.}
	\label{fig:architecture}
	\vspace{-0.15in}
\end{figure}

As seen in Figure \ref{fig:architecture}, the framework consists of 4 primary actors: passengers, goods, vehicles, and the MoD Platform. The MoD platform's main role is to act as an intermediary between customers and vehicles, thereby acting as a marketplace for delivery services. We formulate that the MoD platform is responsible for the following: 
\begin{enumerate}
	\item Matching and Pricing Services: providing the initial matching decisions (based on the proximity of vehicles to ride requests) and facilitate negotiations on pricing. 
	\item Supply/Demand Services: This would involve two subcomponents: (a) maintaining the states such as current locations, current capacity, destinations, etc., for all vehicles. These states are updated in every time step based on the dispatching and matching decisions, and (b) the (Demand Prediction) model used to calculate the future anticipated demand in all zones. 
	\item Location Services: (a) the estimated time of arrival (ETA) model used to calculate and continuously update the estimated arrival time, and (b) routing engine used to generate vehicle’s optimal trajectory to reach a destination.
\end{enumerate} 

{Demand Prediction has been widely studied in literature (e.g., \cite{oda2018movi}, \cite{KE2021102858}) we adopt a zone-based demand prediction approach as in \cite{oda2018movi}.} Since they play a crucial role in enabling this framework, we provide details on the Demand Prediction Model and ETA Model in Appendix B. 

Each of these services exists to communicate information to customers and vehicles' respective decision-making tools. Similar to a real-world scenario, the workflow involves either the passenger or goods customer requesting the MoD platform for a pickup. This initiates an initial matching which is done greedily by the MoD platform's Matching Service. The Vehicle then receives the request and processes it depending on whether it is a passenger or a good. If assigned a passenger, the system facilitates a pricing decision between the vehicles and passengers based on their respective preferences.  Please note that goods are priced using the initial pricing and do not negotiate.  On the passenger's end, we model a utility function to indicate the preferences and accept/reject proposed prices. Goods, on the other hand, follow the initial pricing set by the MoD platform. On the vehicle end, we propose a distributed decision-making algorithm that scales with the increasing number of vehicles. Leveraging the distributed compute in the vehicles, each vehicle is equipped with the following policies: 
\begin{enumerate*}
	\item Route Planning and Matching Policy,
	\item Pricing Policy, and 
	\item DQN Dispatch Policy
\end{enumerate*}.
In the vehicle model, each of these policies exchange information with the MoD Platform as well as amongst themselves to make optimal decisions on:   
\begin{enumerate*}
	\item Which area of the map to dispatch to, 
	\item How much to charge the passenger upon initial matching, and 
	\item Which routes to take in delivering passengers
\end{enumerate*}. 

This framework involves both passengers and agents in the decision-making process. They learn the best pricing actions based on their utility functions that dynamically change based on each agent’s set of preferences and environmental variables.  With a DQN dispatch policy, vehicles act as agents that learn the optimal policy in a distributed manner (at each vehicle) through interaction with the environment. We shall refer to vehicles as agents from this point forth.  Vehicles learn the best future dispatch action to take at time step t, taking into consideration the locations of all other nearby vehicles, but without anticipating their future decisions. Note that, vehicles get dispatched to areas of anticipated high-demand either when they just enter the market, or when they spend a long time being idle (searching for a ride). Vehicles’ dispatch decisions are made in parallel. 
Therefore, our algorithm learns the optimal policy for each agent independently as opposed to centralized-based approaches such as in \cite{oda2018movi}. 

\vspace{-0.1in}
\subsection{Objectives}
In this section, we detail our system's global reward objective which allows efficient fleet dispatch in fulfilling service workloads of different kinds. This global reward is optimized by our proposed algorithm in a distributed fashion as vehicles solve their own DQN (Deep Q-Network) to maximize rewards.

The goals of the system objective include:
\begin{enumerate*}
	\item satisfying the demand of pick-up orders, thereby minimizing the demand-supply mismatch.
	\item minimizing the time taken to pick-up an order (aka pick-up wait time) in tandem to the dispatch time taken for a vehicle to move to a pick-up location.
	\item minimizing the additional travel time incurred by orders due to participating in a shared vehicle.
	\item minimizing the number of vehicles deployed to minimize fuel consumption and traffic congestion, and 
	\item maximizing vehicle earnings.
\end{enumerate*}
\\
This overall objective is optimized at each vehicle in the distributed transportation network. This optimization includes a route planning and matching policy, a pricing policy, and a dispatching policy, all working in tandem with each other. 

\vspace{-0.1in}
\subsection{Multi-hop Scenario}
\begin{figure}
	\includegraphics[width=7.7cm]{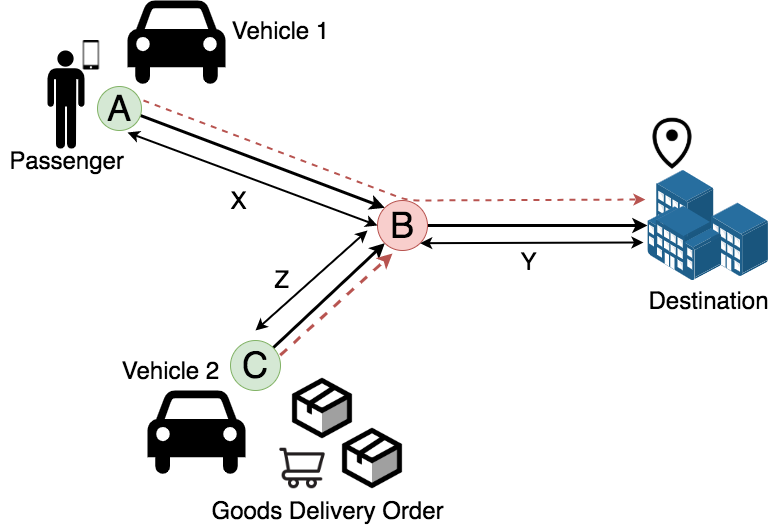}
	\caption{An illustration of transportation scenario with hybrid delivery workload. Vehicle 1 and 2 are in zones A and C respectively. Zone B represents a hop-zone' where the packages part of the goods order packages transfer over to another vehicle}
	\label{fig:MHGD}
	\vspace{-0.15in}
\end{figure}

In Figure \ref{fig:MHGD}, a goods delivery request is to go from zone C to the destination. The goods package can be transported by one vehicle from C to B, and another vehicle from B to the destination. Zone B will serve as a transit location referred to, in this paper, as ``hop-zone''.  This flexibility of changing vehicles is a multi-hop transport of goods. Such multi-hop flexibility improves the packing of goods, as was shown for the case of passengers in \cite{singh2019distributed}. For passenger transfers, it was shown that the multi-hop transfers leads to 30\% lower cost and 20\% more efficient utilization of fleets, as compared to the ride-sharing algorithms. Even though such transfers may not be convenient for passengers, they can be used for goods which motivates the choice in this paper. In addition to the multi-hop transfer of the goods, pooling capabilities of passengers are also exploited for efficient joint transportation system.  

\vspace{-0.1in}
\subsection{Model Parameters and Notations}
In this section,{we describe the notations used to represent, the state, action, and reward spaces for our DQN dispatching algorithm}.The optimization of the system is achieved over $T$ time steps with each step of length $\Delta t$. The fleet make decisions on where on the map to go to serve at each time step $\tau = t_{0},t_{0}+\Delta t,t_{0}+2(\Delta t),\ldots,t_{0}+T(\Delta t)$ where $t_{0}$ is the start time.
The map is split up into a grid with each square being taken as a  zone. Zones are represented by $i \in  \{1,2,3,\ldots,M\}$.  The number of vehicles in the fleet is represented by $N$. A vehicle is marked as \textit{available} if there is remaining seating or trunk capacity.  Vehicles that are completely full or are not considering taking passengers or goods are marked \textit{unavailable}. Available vehicles in zone $i$ at time slot $t$ is denoted $v_{t,i}$. Only available vehicles are eligible to be dispatched. 

{{\bf Vehicles' States: }} The vector \textbf{$X_t$} tracks the vehicle states. For a given vehicle $V$ we denote seating capacity and trunk space for packages as \textbf{$V_{C_{seat}}$} and \textbf{$V_{C_{trunk}}$}, respectively. As a result \textbf{$X_t$} will track: 
\begin{enumerate*}
	\item current zone for vehicle $V$,
	\item available seats,
	\item available trunk space, 
	\item time at which delivery order is picked up, and 
	\item destination zone of each delivery order.
\end{enumerate*} 

{{\bf Demand: }} Pick-up requests at a given zone $i$ at time slot $t$, represents the demand at a given area at that time and is denoted  $d_{t,i}$. The pick-up request demand in each zone is predicted for, $T$ steps ahead, through a historical distribution of trips across the zones, and is denoted $D_{t:T} = (\boldsymbol{\overline{d}_{t}},\ldots,\boldsymbol{\overline{d}_{t+T}})$. 

{{\bf Supply:}} At each time slot $t$, the supply of vehicles for each zone is projected to future time  $\tilde{t}$. Consequently, the number of vehicles that will become available at  $\tilde{t}$ is denoted by $\delta_{t,\tilde{t},i}$. This value is ascertained from an ETA (estimated time of arrival) prediction for all moving vehicles. Consequently, given a set of dispatch actions, we are able to predict the number of vehicles in each zone for $T$ time slots ahead, denoted by $V_{t:T}$.  All the data is combined to represent the environment state space $s_{t}$ by the tuple $(X_{t}, V_{t:T}, D_{t:T})$.
At each assignment of a request to vehicle, the state space tuple is updated with the expected pick-up time, source, and destination data.

\vspace{-0.1in}
\subsection{{Proposed Framework}}\label{Proposed_Architecture}

In addition to the distributed architecture, our key novelties are demonstrated in the policies used for decision making at each agent.  The Route Planning and Matching policy which uses Multi-hop routing, Pricing policy, and DQN Dispatch policy interact with each other and with the MoD platform to optimize for the system objectives. At the agent, the ride requests are input to the system along with the heat map for the supply and demand service of the MoD platform.  It is to be noted that our framework assumes passenger requests are constrained by seating capacity and goods delivery requests are constrained by the vehicle's trunk capacity. 

After having performed an initial greedy vehicle-passenger(s) matching where one request (or more) gets assigned to the nearest vehicle based on its maximum capacity, the MoD platform initiates an initial price for the service to be provided to the assigned customer. Subsequently, the customer and price are communicated to a given agent for planning and pricing. Each agent having received the list of initial matchings (requests) assigned to it with their corresponding initial prices, performs a hop-zone assignment and an insertion operation to plan its route. In this step, vehicles reach their final matchings list by dealing with their initial matchings list in the order of their proximity to it, performing an insertion operation to its current route plan (as long as this insertion satisfies the capacity, extra waiting time, and additional travel distance constraints to guarantee that serving this request would yield a profit). 

The vehicles adopt a dispatching policy using DQN, where they get dispatched to zones with anticipated high demand when they experience large idle duration or when they newly enter the market. If the customer is a passenger, using the expected discounted reward learned from DQN and the ride’s destination, vehicles weigh their utility based on the potential hotspot locations and propose a new price for the customer. This takes place on a customer-by-customer basis, where a vehicle upon inserting a customer to its current route plan, proposes to him/her the new price. Then, the customer has the ability to accept or reject based on his/her own independent utility function. Finally, the vehicle upon receiving the customer’s decision, either confirms the addition of this customer to its route plan or removes him/her. A Vehicle communicates with the MoD platform, as needed, to request new information on other vehicles (prior to making a dispatch or price decision) or update its own status (after any decision). 

\begin{algorithm}
	\caption{Framework Algorithm}\label{alg:main}
	\begin{algorithmic}[1]
		\State \textbf{Initialize} vehicles' states $X_0$ and $t_0$

		\For {$t \in T$}
			\State \textbf{Fetch} all vehicles that entered the martket in time slot t, $V_{new}$
			\State \textbf{Dispatch} $V_{new}$ to zones with anticipated high demand using Algo. \ref{alg:dispatch_algo}
			\State \textbf{Fetch} all ride requests at time slot t, $D_{t}$
			\State \textbf{Fetch} all available vehicles at time slot t, $V_{t}$
			\For{ each vehicle $V_j \in V_t$}
				\State \textbf{Obtain} initial matching $A_j$ using Algo. \ref{alg:greedy_matching}
				\For {each request $r_i \in A_j$}
					\If {request is passenger} :
							\If {vehicle seating capacity $V^j_{C_{seat}} == 0$}
							\State \textbf{Insert} $r_i$ to $D_{t+1}$
							\State \textbf{Continue}
							\EndIf
						\State \textbf{Obtain} initial pricing $P_{init,pass}$ using Eq. (\ref{eqn:init_price})
						\State \textbf{Perform} Route planning using Algorithm \ref{alg:route_planning}
						\State \textbf{Obtain} $S'_{V_{j}}[r_{i}]$ based on $cost(V_j, S'_{V_{j}}[r_{i}])$
						\State \textbf{Update} trip time $T_i$ based on $S'_{V_{j}}[r_{i}]$ using ETA model
						\State \textbf{Calculate} final price $P(r_i)$ based on $S'_{V_{j}}[r_{i}]$ using Eq. \ref{eqn:update_price}
						\State \textbf{Get} customer $i$'s final decision $C^i_d$ on $P(r_i)$ using Eq. \ref{eqn:passenger_utility} and Eq. \ref{eqn:passenger_decision}
						\If {$C^i_d == 1$}
							\State \textbf{Update} $S_{V_j}$ to $S'_{V_{j}}[r_{i}]$
						\Else
							\State \textbf{Insert} $r_i$ to $D_{t+1}$
						\EndIf
					\ElsIf {request is good}:
						\If {vehicle trunk capacity $V^j_{C_{trunk}} == 0$}
						\State \textbf{Insert} $r_i$ to $D_{t+1}$
						\State \textbf{Break}
						\EndIf
						\If {vehicle capacity $V^j_{C} == 0$} 
							\State \textbf{Obtain} price using $P_{init,good}$ using Equation (\ref{eqn:init_price})
							\State \textbf{Perform} Route planning using Algorithm \ref{alg:route_planning}
							\State \textbf{Obtain} $S'_{V_{j}}[r_{i}]$ based on $cost(V_j, S'_{V_{j}}[r_{i}])$
							\State \textbf{Update} trip time $T_i$ based on $S'_{V_{j}}[r_{i}]$ (ETA model)
						\Else
							\State \textbf{Assign} next destination using Algorithm \ref{alg:hopzone_assignment}
							\State \textbf{Perform} Route planning using Algorithm \ref{alg:route_planning}
							\State \textbf{Obtain} $S'_{V_{j}}[r_{i}]$ based on $cost(V_j, S'_{V_{j}}[r_{i}])$
							\State \textbf{Update} trip time $T_i$ based on $S'_{V_{j}}[r_{i}]$ (ETA model)
						\EndIf
					\EndIf					
					\State \textbf{Update} State Vector $s_t$
				
				\EndFor
				\State \textbf{Retrieve} next stop from $S_{V_j}$
				\State \textbf{Drive} to next stop 
			\EndFor
			\State \textbf{Fetch} all idle vehicles with Idle\_duration $>$ 10 minutes, $V_{idle}$
			\State \textbf{Dispatch} $V_{idle}$ to zones with anticipated high demand using Algo. \ref{alg:dispatch_algo}
			\State \textbf{Update} State Vector $s_t$
		\EndFor
			
	\end{algorithmic}
\end{algorithm}

We present the proposed framework's logic in Algorithm \ref{alg:main}. In line 1, we initialize the states of the vehicles and consequently run the system at each time step which may be discretized to $t$. In lines 3-4, the MoD platform initializes the vehicle marketplace and initializes their locations using the dispatching algorithm shown in Algorithm \ref{alg:dispatch_algo}. The MoD platform then fetches all requests and currently available vehicles in lines 5-6 respectively. It is to be noted that the iterations across each vehicle shown in lines 7-40 may be run asynchronously at each vehicle. Each vehicle is therefore able to run the services indicated on the Vehicle Platform (see figure \ref{fig:architecture}). In line 8, the MoD greedily assigns the nearest set of requests $A_j$ to each vehicle using Algorithm \ref{alg:greedy_matching}. For each of these requests, we determine if it is a passenger or goods as in lines 10 and 24, respectively. If a passenger has been assigned, the vehicle retrieves the initial pricing which is determined by Equation (\ref{eqn:init_price}) as seen on line 14. Lines 15-17 perform insertion-cost based route planning as in Algorithm \ref{alg:route_planning}. The pricing strategy is executed on line 18 and the accept/reject decision is obtained in line 19 based on the utility function of the passenger. In the event that a goods package has been assigned, we obtain an initial price based on Equation (\ref{eqn:init_price}) as shown on line 29. We then perform multi-hop assignment in line 30 based on Algorithm \ref{alg:hopzone_assignment} followed by insertion-cost-based route planning on line 30. The two combined yield our novel multi-hop route planning for goods. It is to be noted that we place goods in the trunk capacity of a given vehicle, and the passengers in the seating capacity. 

As we see in the overall Algorithm, the key components include the route planning and matching policy, pricing policy, and the dispatch policy. These will be discussed in the following sections.

\section{Matching and Route Planning Policy}\label{Routing}

The vehicle-request assignment task, which we refer to as Matching, is performed for both passengers and goods. A passenger request is initially matched by the MoD platform before insertion cost based route planning by the vehicle and a pricing agreement (described in Section \ref{Pricing}). A goods delivery request, on the other hand, is also initially matched by the MoD platform before dynamic multi-hop and insertion cost based route planning. In both cases, the overall matching policy consists of the sequences of events described below.

\vspace{-0.1in}
\subsection{Initial Matching}
The MoD platform initially performs a greedy vehicle-request assignment using Algorithm \ref{alg:greedy_matching}. In this procedure, each request is allocated to the nearest vehicle that minimizes waiting time given all the vehicles' capacity constraints and available status. It is important to note that passengers are allocated to the seating capacity while goods are allocated to the trunk capacity.

In Algorithm \ref{alg:greedy_matching}, we input the set of available vehicles on the map $V_t$ and the demand backlog $D_t$. The algorithm ultimately outputs a set of request-to-vehicle assignments $A_j$. In Line 3, we initialize an empty set of assignments for $A_j$ and fetch the capacities of each of the available vehicles into $V^j_{capacity}$. Consequently after line 4, for each request $r_i$ in the demand backlog $D_t$, we first check whether $r_i$ is a passenger or good and fetch vehicles seating capacity $C_{seat}$ or trunk capacity $C_{trunk}$ accordingly (see lines 5-6). Upon obtaining the location of vehicles in line 7, we calculate the trip time in line 8 and select the vehicle $V_j$ that minimizes the waiting time for request $r_i$. We then append the assignment to $A_j$.

Given the set of initial assignments $A_j$, we proceed to the next steps depending on if an assignment is for a passenger or a good. 

\begin{algorithm}
	\caption{Greedy Initialization}\label{alg:greedy_matching}
	\begin{algorithmic}[1]
		\State \textbf{Inputs}: Available Vehicles $V_{t}$, Requests $D_{t}$ including origins $o$ and destinations $d$
		\State \textbf{Outputs}: Request to Vehicle Assignments $A_j$ for each $V_j \in V_t$
		\State \textbf{Initialize} $A_j = []$, $V^j_{capacity} = V^j_C$ for each $V_j \in V_t$
		\For {each $r_i \in D_{t}$}
			\If {$r_i$ is a passenger}
				$C = C_{seats}$
			\ElsIf{$r_i$ is a good}
				$C = C_{trunk}$
			\EndIf
			\State \textbf{Obtain} locations of candidate vehicles $V_{cand}$, such that: $\lvert loc(V_j) - o_i \rvert \leq  5 km^{2} \textbf{and} (V^j_{C} + \lvert r_i \rvert)  \leq C^{V_j}_{max}$.
			\State \textbf{Calculate} trip time $T_{j,i} \in t_{cand,i}$ from $loc(V_j) \in V_{cand}$ to $o_i$ using the ETA model. 
			\State \textbf{Pick} $V_j$ whose $T_{j,i} = argmin(T_{cand.i})$ to serve request $r_i$
			\State \textbf{Push} $r_i$ to $A_j$
			\State \textbf{Update} $loc(V_j) \gets o_i$
			\State \textbf{Increment} $V^j_{capacity} \gets V^j_{capacity} + \lvert r_i \rvert $
		\EndFor
		\State \textbf{Return} $A_t = [A_j, A_{j+1}, ..., A_n]$, where $n=\lvert V_t \rvert$
	\end{algorithmic}
\end{algorithm}

\vspace{-0.1in}
\subsection{Hop-zone Assignment for Goods}

In transporting goods, we present a route planning framework that is analogous to hitch-hiking. A package is routed through layovers or ``hop-zones" (as we will refer to henceforth in this paper), in an iterative fashion such that at each iteration, the system maximizes the distance towards the final destination. This way, at each leg of the package's journey, we allocate the locally optimal hopzone which is explained in Algorithm \ref{alg:hopzone_assignment}. We refer to this as Multi-hop routing. 

To facilitate hop-zones in practice, we consider the suggestion from \cite{crowddeliver} where each hop-zone is a public point of interest such as a store, post office, or fuel station. We assume incentives for these locations to hold packages that are in transit along their journey to the destination. 

Algorithm \ref{alg:hopzone_assignment} determines which hop-zone to allocate for a given goods request $r_i$. Given an input of the location information contained in $r_i$, vehicle $V_j$'s current route $S_{V_j}$, a drop-off threshold radius $d_{drop}$, and a minimum distance gain tolerance $d_{gain_{min}}$, the Hop-zone assignment algorithm outputs the ``next-destination" for $r_i$. 

$d_{drop}$ and $d_{gain_{min}}$ are two hyper parameters that suggest the minimum threshold below which a hop-zone transit is no more desirable. As seen in line 5, if the request is already within the drop-off radius, the next destination assigned is the final destination of the request. 

Given the vehicle's current route $S_{V_j}$, lines 10-14 first check for the stop that maximizes the distance towards the final destination of the request. Upon finding the most suitable stop, lines 16-21 find the nearest hopzone to the stop. This is deemed the most suitable hop-zone unless the threshold parameters are not met as checked in lines 22,23.  

\begin{algorithm}
	\caption{Hop-zone Assignment}\label{alg:hopzone_assignment}
	\begin{algorithmic}[1]
		\State \textbf{Inputs}: Goods Request (Origin $o_i$, Current Location $l_i$, Final Destination $d_i$), Route Plan $S_{V_j}$, Drop-off Radius $d_{drop}$, Min Distance Gain $d_{gain_{min}}$ 
		\State \textbf{Outputs}: Next Destination $h_i$
		
		\State \textbf{Fetch} All available hop-zone locations $H$
		\State \textbf{Compute} distance from Current Location to Final Destination  $d_0 = dist(l_i, d_i)$ 
		\If {$d_0 < d_{drop}$}
			\State \textbf{Return} $d_i$
		\Else
			\State \textbf{Initialize} $h_i$ as $S_{V_j}[0]$
			\State \textbf{Initialize} distance $d_s = \infty$ 
			\For {stop $s \in S_{V_j}$}
				\State \textbf{Compute} distance $d_s'$ such that $d_s' = dist(s, d_i)$
				\If {$d_{s'} < d_s$} 
					\State \textbf{Update} $d_s \gets d_{s'}$
					\State \textbf{Update} $h_i \gets s$
				\EndIf
			\EndFor
			\State \textbf{Initialize} distance $d_h = \infty$ 
			\For {hop-zone $h \in H$}
				\State \textbf{Compute} distance $d_h'$ such that $d_h' = dist(h, h_i)$
				\If {$d_{h'} < d_s$} 
					\State \textbf{Update} $d_h \gets d_{h'}$
					\State \textbf{Update} $h_i \gets h$
					\State \textbf{Compute} distance gain $d_{gain} = (dist(h, h_i) - d_0)/d_0$
				\EndIf
			\EndFor
			\If {$d_h < d_{drop} |  d_{gain} < d_{gain_{min}}$} 
				\State \textbf{Return} $d_i$
			\Else 
				\State \textbf{Return} $h_i$
			\EndIf
		\EndIf

		\State \textbf{Return} Hop-trips
	\end{algorithmic}
\end{algorithm}

\vspace{-0.1in}
\subsection{DARM Distributed Optimization for Route Planning}
We use an  insertion-cost based route planning for both passenger and goods delivery requests, extending the approach in \cite{haliem2020distributed}. If the request is a good, it proceeds to this DARM framework for route planning after hop-zone assignment from Algorithm \ref{alg:hopzone_assignment}.   If the request is a passenger, DARM route planning is performed immediately after initial assignment from Algorithm \ref{alg:greedy_matching}. The details are provided in Appendix \ref{apd:darm}.

\vspace{-0.2in}
\section{Distributed Pricing Policy} \label{Pricing}

In this section, we will discuss the components for pricing based strategy, that is part of the distributed ride-sharing framework detailed in Section \ref{System}. When a customer requests a pickup, the MoD platform assigns a vehicle and proposes an initial price for the journey. For this initial price, we build upon what has been proposed in DPRS (Distributed pricing based ride-sharing) \cite{haliem2020distributed}. In our simulated environment, we considered various vehicle types: hatch-back, sedan, luxury, and van. Each of which have varying seating capacities, trunk capacities, mileage, and base price for vehicle $j$ per trip denoted by $B_j$.

Given that we consider passengers and goods/packages, it is to be noted that the pricing policy from the perspective of the driver is based on which type of customer has been assigned. For each delivery of goods, vehicles adhere to the initial pricing provided by the MoD platform. On the other hand, the vehicles follow the negotiation-based pricing strategy proposed by DPRS \cite{haliem2020distributed} when picking up passengers. In such a case, each passenger expresses their preferences through a defined utility function in the negotiation process. Likewise, vehicles express their preferences based on their anticipated earnings in picking up the passengers. These steps are discussed in this section. 

\vspace{-0.1in}
\subsection{Initial Pricing}\label{Initial_Pricing}
When first matched, the MoD platform proposes the initial price taking into consideration the following factors:
\begin{itemize}
	\item \textbf{Trip Distance:} This includes the distance till pickup as well as the distance from pickup to drop off. As shown in Algorithm \ref{alg:route_planning}, this distance is computed using the weights of the $n$ edges that constitute the vehicle’s optimal route from its current location to origin $o_i$ and to destination $d_i$. This route is obtained through the insertion-operation, the route which minimizes the DARM cost function.
	
	\item \textbf{{Customer/Package waiting time:}} The time elapsed from request till pickup for a trip $i$ is denoted as $T_i$.
	
	\item {\textbf{Number of shared entities:} Number of other passengers or goods who share travelling a trip distance. This is determined from the vehicle's route considering whether all or part of the trip is shared. A vehicle $j$ 's capacity is defined as the combined capacity $V^C_j = V^{C_{trunk}}_j  + V^{C_{seats}}_j$. The capacity of vehicle $j$ when it reaches the origin $o_i$ location of this request $r_i$ subtracted from its capacity when it reaches the destination $d_i$ , $V^{C}_{j} [d_i ]$. We therefore define $V^{C}_{j} [r_{i}] = | V^{C}_{j} [d_{i}] - V^{C}_{j} [o_{i}] | $}. 
	
	\item \textbf{Fuel consumption:} The cost for fuel consumption associated with this trip, denoted by $D_i * (P_{gas}/M_j)$, where $P_{gas}$ represents the average gas price, and $M_V^j$ denotes the mileage for vehicle j assigned to trip $i$. { We note that $D_i = S_{V_j}[r_i]$ initially, but then as vehicle $j$'s route gets updated to $S^{\prime}_{V_j}[r_i]$ every time a customer is added to its route,  $D_i$ becomes $S^{\prime}_{V_j}[r_i]$ to calculate the fuel consumption of the updated route. }
\end{itemize}

The overall price initialization equation for request $r_i$ is:
\vspace{-0.1in}
\begin{multline}
P_{init}[r_i] = B_j + \left[ \omega^1 * \frac{cost(V_j, S_{V_j}[r_i])}{V^C_j [r_i]} \right] \\
\text{ } +  \left[ \omega^2 * \left( \frac{cost(V_j, S_{V_j}[r_i])}{V_j^{C_j} [r_i]} \right)  * \frac{P_{gas}}{M^C_j}   \right]  - \left[ \omega^3 * T_i \right] 
\label{eqn:init_price}
\end{multline}

{where $\omega^1$, $\omega^2$, and $\omega^3$ are the weights associated with each of the factors affecting the price calculation. $\omega^1$ is the price per mile distance according to the vehicle type.  $\omega^2$ is set to 1 as it doesn’t change across vehicles, what changes is the mileage in this factor. Finally, $\omega^3$ is the price per waiting minute that is influenced by the vehicle type, it is negative here as we want to minimize the waiting time for the customer.}

{We note that, for ride-sharing customers: the trip distance is calculated as the DARM insertion cost $S_{V_j}[r_i]$ obtained from Algorithm \ref{alg:route_planning}. This initial pricing gets updated for on-board passengers whenever a vehicle picks up an additional customer (as the $V^{C}_{j}$ will now increase), where all the overlapping (shared) distances are taken into consideration (when constructing $S^{\prime}_{V_j}[r_i]$) and thus price may get reduced. Our proposed algorithm will first use the initial price and notify the driver, who will then modify the pricing based on the Q-values of the driver’s dispatch-to location.}

\vspace{-0.1in}
\subsection{Vehicle Proposed Pricing}\label{Vehicle_Proposed_Pricing}
The core intuition behind assessing the cost/benefit of picking up a passenger is in having knowledge over the supply-demand distribution over the city. This supply-demand distribution is learned by each vehicle through the DQN dispatch policy. The intent of the dispatch policy is to provide a given delivery vehicle with the best action to take which is predicted after weighing the expected discounted rewards (Q-values) associated with each possible move on the map using DQN (described in Section VI). {We note that, with the profits term being part of the reward function,  the rewards (Q-values) have embedded information about the travel distance and fuel consumption. Thus, Q-values are sufficient to reflect the profitability of serving at a given zone over the map. With the dispatch policy running every so often (set to 5 minutes in our simulation), the vehicle is able to gain the necessary insight on the supply-demand distribution (as well as all other factors embedded in the reward function) and make informed decisions on the pricing strategy.} This decision making sequence is captured in the following steps: With the knowledge on expected discounted sum of rewards (Q-values) across the map, the vehicle:

\begin{itemize}
\vspace{-0.05in}
	\item Ranks regions on the map by expected discounted sum of rewards, which may be obtained using the DQN's Q values.  We denote this rank as $\alpha$. 
	
	\item Maintain a set of $\lambda$ highest ranked regions as $L$. 
	 
	\item Upon Route-planning, re-compute initial pricing $P_{init}(r_i)$ using the updated route $S'_{V_j}$ using Equation (\ref{eqn:init_price}). This way, the vehicle takes detours required from its current path into account as it serves incoming request $r_i$, thus encouraging requests going in the same direction to be matched together.
	
	\item If $loc(r_i)$ is in $L$, propose $P_{init}(r_i)$, otherwise the vehicle will propose a higher cost to make up for the cost of mobilizing to a region of low demand. In such a case, the proposed price $P(r_i)$ is computed using Eq. (\ref{eqn:update_price}).

\vspace{-0.1in}
\begin{equation}
  P(r_i) = P_{init}(r_i) + [P_{init}(r_i) * \frac{\alpha_{loc(r_i)}}{2} * B_j]
  \label{eqn:update_price}
\end{equation}
\end{itemize}

\vspace{-0.1in}
\subsection{Passenger Decision Function}\label{Passenger_Decision_Function}
Upon a vehicle's proposed price, the passenger may accept or reject the price. This is achieved through the passenger's utility function as follows: 

\vspace{-0.1in}
\begin{equation}
U_i = [\omega^4 * \frac{1}{V^j_C}] + [\omega^5 * \frac{1}{T_i}] + [\omega^6 * \frac{1}{V^j_T}]
\label{eqn:passenger_utility}
\end{equation}

In Equation (\ref{eqn:passenger_utility}) above, we assign weights to the passenger preferences and consider 3 components: {(i) preference in sharing the vehicle expressed by $\omega^4$, this is captured in the utility equation based on the current capacity of vehicle $j$ assigned to trip $i$, denoted by ${V^j_C}$. } (ii) tolerance in waiting time $T_{i}$ expressed by $\omega^5$ (iii) preference in vehicle type $V^{j}_{T}$ expressed by $\omega^6$, {where $\omega^4$, $\omega^5$, and $\omega^6$ are the weights associated with each of the factors affecting the customer’s overall utility. These values are set at random for each customer to represent a variety of customer preferences that can be accommodated by our system.
The decision on whether or not a passenger accepts a proposed price is determined by the value of the utility function. We introduce a flexibility factor  $\delta_i$ to represent how much the customer $i$ is willing to compromise in the decision-making process}. If the value of the utility is $\geq$ the price multiplied by $\delta_i$, then the customer accepts as follows:

\vspace{-0.1in}
\begin{equation}
C^i_d=\begin{cases}
1, & \text{if $U_i \geq P(r_i)\delta_i$}.\\
0, & \text{otherwise}.
\end{cases}
	\label{eqn:passenger_decision}
\end{equation}

\section{Distributed DQN Dispatch Policy}\label{Dispatching}
We utilize a distributed DQN dispatch policy to rebalance idle vehicles to areas of anticipated demand and profit.  We utilize a reinforcement learning framework, with which we can learn the probabilistic dependence between vehicle actions and the reward function thereby optimizing our objective function. At each time step $t$, idle vehicles observe the state of the environment, perform inference on their trained DQN to dispatch to regions on the map where they can anticipate a reward. This ultimately improves fleet utilization. Algorithm \ref{alg:dispatch_algo} describes the dispatching of vehicles by running inference on the trained DQN. Line 5 of the algorithm specifically describes the best action that a given vehicle $V_n$ infers from the trained Q-network given the state $s_{t,n}$ and set of possible actions $a_{t,n}$. 

The fleet of autonomous vehicles was trained in a virtual Spatio-temporal environment that simulates urban traffic and routing. In our simulator, we used the road network of the New York City Metropolitan area along with a realistic simulation of taxi pick-ups and package delivery requests. This simulator hosts each deep reinforcement learning agent which acts as a delivery vehicle in the New York City area that is looking to maximize its reward (Eq. \eqref{eq:DQNreward}). In our algorithm, we define the reward $r_{t,n}$ as a weighted sum of different performance components that reflect the objectives of our DQN agent. We present the state, action, and reward for the dispatch policy: \\
\textbf{\textit{State}}: The state variables defined in this framework capture the environment status and thus influence the reward feedback to the agents' actions. We discretize the map of our urban area into a grid of length $\boldsymbol{x}$ and height $\boldsymbol{y}$; resulting in a total of $\boldsymbol{x}*\boldsymbol{y}$ zones. This discretization prevents our state and action space from exploding thereby making implementation feasible. The state at time $t$ is captured by following tuple: $(X_{t},V_{t:T} , D_{t:T})$. These elements are combined and represented in one vector denoted as $s_{t}$. When a set of new ride requests are generated, the simulator engine updates its own data to keep track of the environment status. {Since the state space is large, we don’t use the full representation of $s_{t}$, instead a map-based input is used to alleviate this massive computing}. The three-tuple state variables in $s_{t}$ are passed as an input to the DQN input layer which consequently outputs the best action to be taken.
\begin{enumerate}
	\item 
	\textbf{$X_t$} will track vehicle seating capacity and trunk space for goods/packages. $V_{C_{seat}}$ and $V_{C_{trunk}}$ respectively. As a result \textbf{$X_t$} will track: \textit{current zone of vehicle $v$, available seats, available trunk space, pick-up time of delivery order, destination zone of each order}. 
	\item
	\textbf{$V_{t:T}$} is a prediction of number of available vehicles at each zone for $T$ time slots ahead. 
	\item
	\textbf{$D_{t:T}$} has a term \textbf{$d_{t:T}$} that predicts joint demand of passengers \& goods delivery orders at each zone for $T$ time slots ahead. 
\end{enumerate}

\textbf{\textit{Action}}: The action of vehicle $n$ is denoted by  $a_{t,n}$  decides the zone it should be dispatched to at time slot t, which is given as $Z_{t,i}$.  {In our simulator, the vehicle can move (vertically or horizontally) at most 7 cells, and hence the action space is limited to these cells. A vehicle can move to any of the 14 vertical (7 up and 7 down) and 14 horizontal (7 left and 7 right). This results in a 15x15 action space $a_{t, n}$ for each vehicle as a vehicle can move to any of these cells or it can remain in its own cell. }
If vehicle $n$ is full it can not serve any additional customer. Alternatively, if a vehicle decides to serve current customers, the shortest route is used along the road network to pickup the assigned orders.

\textbf{\textit{Reward}}: 
The reward function which drives the dispatch policy learner's objectives, is shaped in a manner which (1) minimize the supply-demand mismatch for passengers and goods: (${diff}_t$), (2) minimize the dispatch time: $T_t^D$ (i.e., the expected travel time of vehicle $V_j$ to go zone m at time step $t$), (3) minimize the extra travel time a vehicle takes for car-pooling compared to serving one customer:$\Delta_t$, (4) maximize the fleet profits $P_t$, and (5) minimize the number of utilized vehicles: $e_t$. 
Below is the reward function $r_{t,n} = r(s_{t,n} , a_{t,n})$ which is used at each agent $n$ of the distributed system at time $t$. The weights shown in Equation (\ref{eq:DQNreward}) are used only in instances when an agent is not fully occupied and is eligible to pickup additional orders. 

\vspace{-0.3in}
\begin{multline}
r_{t,n} = \beta_1(b_{t,n} + p_{t,n}) - \beta_2c_{t,n} - \beta_3 \sum_{u=1}^{U_n} \delta_{t,n,u} \\ 
\text{ } -\beta_4 \mathbb{P}_{t,n} + \beta_5 \max (e_{t,n} - e_{t-1,n}, 0) 
\label{eq:DQNreward}
\end{multline}

The reward function  as a linear combination of the following terms along with their respective weights $\beta_i$: (1) total number of customer orders picked up $b_{t,n}$ denotes the number of passengers served by vehicle $n$ at time $t$ and $p_{t,n}$ denotes the number of packages being carried in the trunk of vehicle $n$ at time $t$. (2) $c_{t,n}$ denotes the time taken by vehicle $n$ at time $t$ to hop or take detours to pick up extra orders. This term discourages the agent from picking up additional orders without considering the delay in current passengers/goods orders. (3) $\sum_{u=1}^{U_n} \omega_u \cdot \delta_{t,n,u}$ denotes the sum of additional time vehicle $n$ is incurring at time $t$ to serve additional passengers or packages. (4) $\mathbb{P}_{t,n}$ denotes the profit, {which is calculated by subtracting cost of fuel consumption from revenue}. (5) $\max (e_{t,n} - e_{t-1,n}, 0)$ addresses the objective of minimizing the the number of vehicles at $t$ to improve vehicle utilization.

While the primary role of the DQN is to act as a means of dispatching idle vehicles, it contains useful signals on future anticipated demand that is utilized by other components of our method including the Demand Aware Route Planning (DARM), and Pricing Policy for Passengers. The profits component $P_t$ of the reward function encodes the expected discounted rewards (Q-values) corresponding to each position on the map with the expected earnings at these locations. As a result, vehicles now have a readily available representation of the supply-demand distribution of the city. This representation is essential in informing decisions for route planning and pricing proposals which are described in Section \ref{Routing} and Section \ref{Pricing} respectively. 
\begin{algorithm}[t]

	\caption{Dispatching using DQN}
	\label{alg:dispatch_algo}
	\begin{algorithmic}[1]
		\State \textbf{Input:}  $X_t, V_{t:T}, D_{t:T}$
		\State \textbf{Output:} Dispatch Locations
		\For {each vehicle}
			\State \textbf{Construct} a state vector $s_{t,n} = (X_t, V_{t:T}, D_{t:T})$ 
			\State \textbf{Push} state vector to Q-Network 
			\State \textbf{Get} best action $a_{t,n} = argmax[Q(s_{t,n},a,\theta)]$ which represents dispatch zone 
			\State \textbf{Update} Dispatch Locations for vehicle
		\EndFor
		\State \textbf{Return} Dispatch Locations
	\end{algorithmic}

\end{algorithm}

\section{Experiments}\label{Experiments}
 In this section, we describe the methodology of our experimentation and detail the evaluation of the proposed framework against baselines. The overall objective of our experiments was to observe the key performance metrics of a fleet in an environment that is representative of a real-world use case. 

\vspace{-0.1in}
\subsection{Methodology}
\subsubsection{Simulator}
Our simulator set up in the following experiments mimicked the MoD architecture as illustrated in Figure \ref{fig:architecture}. The customer, MoD Platform, and vehicles were implemented along with the services and policies specified.  We developed a discrete event simulator using Python to spawn customers based on a realistic passenger and goods request dataset which is described in detail in Appendix A. Specifically, we consider the New York Metropolitan area and utilize the Open Source Routing Engine (OSRM) as the backend for the Routing Engine of the MoD platform.  For each trip, we obtain the pick-up time, passenger count, origin location, and drop-off location. We use this trip information to construct travel requests demand prediction model as well. 

We initialize the simulation by populating vehicles over the map constrained around the major boroughs of New York City. Given we are considering different vehicle types, we randomly assign a type and initial location to each vehicle. Each vehicle type is assigned characteristic features as follows: mileage, maximum seating capacity, maximum trunk capacity, and price rates (per mile of travel distance $\omega^1$, and per minute of waiting $\omega^2$). {We set the number of fleet vehicles to 15000 to serve the passenger and goods workload. It is to be noted that not all vehicles are populated at once. Vehicles are deployed incrementally into the market at each time step $t$.} 

We also defined a reject radius threshold for a customer request. Specifically, when a request is created by a customer and there is no vehicle within a radius of 5km, it is rejected. 

\subsubsection{Multi-Hop}

Our simulator was configured with a total of 222 hop-zones evenly distributed accross the New York metropolitan area. Each hopzone was initialized with a maximum holding capacity of 1000 packages. These hop-zone locations are provided to Algorithm \ref{alg:hopzone_assignment} to be considered in routing for goods. 

For our experiments, we set the drop-off radius $d_{drop} = 2km$. Additionally, we defined a minimum distance gain threshold to be $d_{gain_{min}} = 0$. This way, the algorithm is able to avoid assigning hop-zones which take the package further away from its destination. 

\subsubsection{DQN Training and Testing}

We use the data of June 2016 for training, and one week from July 2016 for evaluations. We trained our DQN neural networks using the data from June 2016 for 10000 epochs and used the most recent 5000 experiences as a replay memory.

For each experiment, we ran training using data from the month of May for a total of 14 days. Upon saving Q-network weights, we retrieve the weights to run testing on an additional 7 days from the month of June. 

\subsubsection{Evaluation Metrics}

Tying back to the objective of our system, an effective algorithm is able to perform well in maximizing the number of customers delivered, maximize the profits for vehicles, minimize the idle driving time (thereby improving fleet utilization), and minimize the customer waiting time. Taking these into consideration, for each of the considered algorithms, we evaluate the following metrics to observe the performance of the fleet as a whole. 
\begin{itemize}[leftmargin=*]
	\item \textbf{Accepted Requests}: The total number of customers served indicates how effectively the algorithm is able to minimize the supply demand gap and fulfill delivery requests. {We investigate both the accepted requests for goods and passengers separately}.
	\item \textbf{Rejected Requests}: The total number of customers rejected.  {We also show this metric for rejected goods' requests and rejected passengers' requests separately}.
	\item \textbf{Occupied Vehicles}: The total number of vehicles currently carrying passengers or goods. {We further show the average occupied seating capacity by passengers and the average occupied trunk capacity by goods.}
	\item {\textbf{Waiting Time: } This captures the time that passengers or goods had to wait to be picked up. We note that wait time is an important metric for customer convenience with mobility-on-demand services. So, we show this for passengers and goods separately.}
	\item \textbf{Profits}: This is the average amount of profits accumulated by a vehicle over the course of a day.
	\item \textbf{Cruising Time}: This represents the time at which a vehicle is neither occupied nor gaining profit but still incurring gasoline cost. Lower cruising times therefore suggests a cost effective policy. 
	\item \textbf{Travel Distance}: The total amount of distance during which fuel was being consumed.  
	\item \textbf{Occupancy Rate}: This is defined as the percentage of time where vehicles are occupied out of their total working time.
\end{itemize}

\begin{figure*}
	\centering
	\begin{minipage}[t]{0.8\textwidth}
		\resizebox{\textwidth}{!}{  
			\includegraphics[trim = 50 0 50 50]{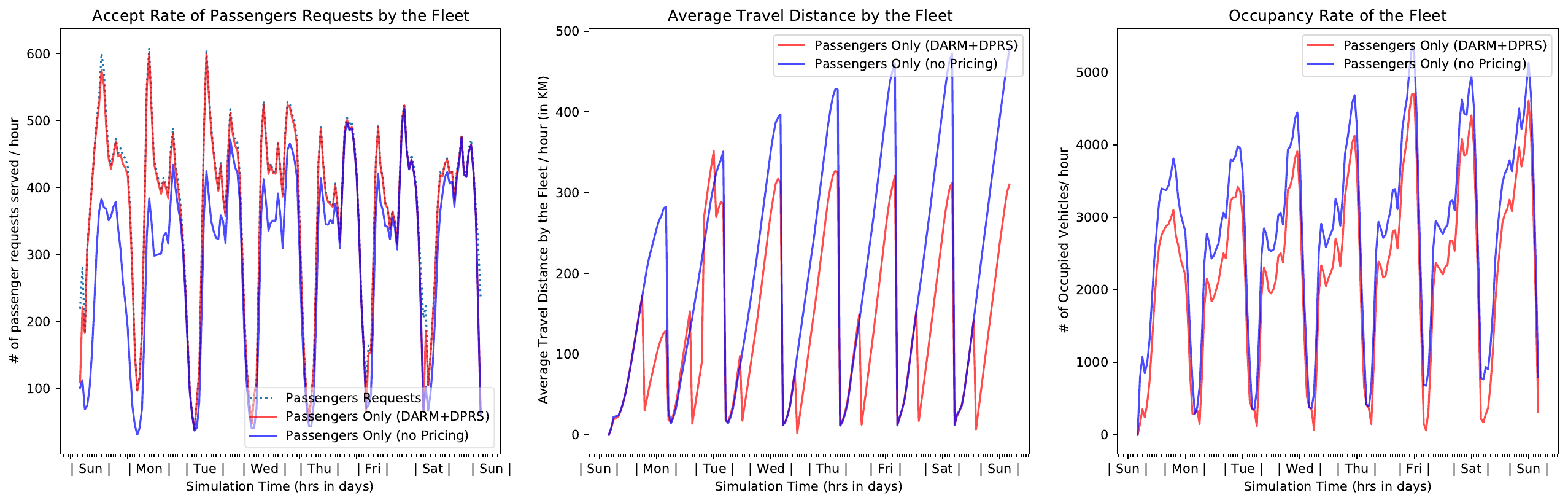}}
		\caption{\small Performance metrics for an Independent Load of passengers with DPRS Vs.  a baseline with no pricing}\label{fig:metrics_DPRS}
	\end{minipage}
	\hspace{.01in}
	\begin{minipage}[t]{0.8\textwidth}
		\resizebox{\textwidth}{!}{  
			\includegraphics[trim = 50 0 50 0]{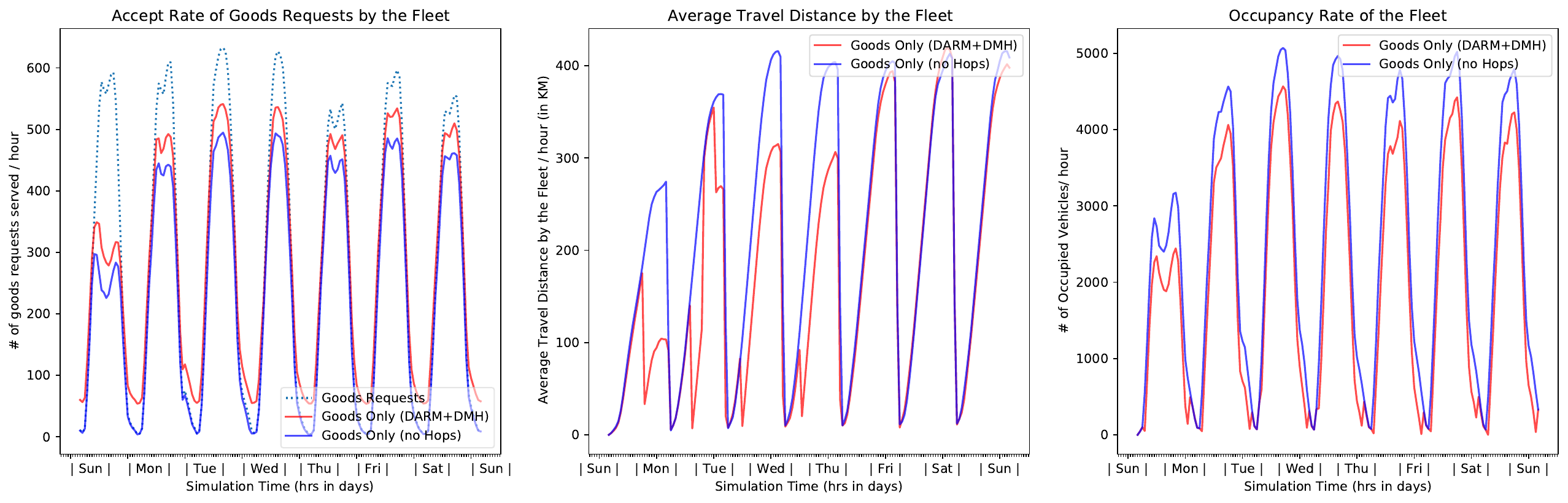}}
		\caption{\small Performance metrics for an Independent Load of Goods with DMH Vs.  a baseline with no multi-hops}\label{fig:metrics_hops}
	\end{minipage}
\end{figure*}
\vspace{-0.1in}
\subsection{Baselines}
Given our proposed novelties of multi-hop route planning, and combined goods+passenger workload transportation, we formulate the following baselines to evaluate the performance of the novelties relative to existing methods:
\begin{enumerate}[leftmargin=*]
	\item \textit{Combined Load (DARM + DPRS + DMH)}: This is our proposed method where we extend DARM + DPRS with Dynamic Multi-hop (DMH) route planning for goods delivery. Here we consider a combined load where each vehicle transports a combination of goods and passengers.
	\item \textit{Combined Load (DARM + DPRS)}: This is the baseline with an insertion cost route planning and matching along with distributed pricing policy. Here we consider a combined load where each vehicle transports a combination of goods and passengers. Note that Dynamic Multi-hop routing has been omitted.
	\item \textit{Independent Load (DARM + DPRS + DMH)}: This is the baseline with an insertion cost and multi-hop route planning and matching along with distributed pricing policy. Here we consider an independent load where each vehicle transports either goods or passengers exclusively.
	\item \textit{Independent Load (DARM + DPRS)}: This is the baseline with an insertion cost route planning and matching along with distributed pricing policy as in \cite{haliem2020distributed}. Here we consider an independent load where each vehicle transports either goods or passengers exclusively.  Note that Dynamic Multi-hop routing has been omitted.
\end{enumerate}

The proposed baselines aim to evaluate the effectiveness of our dynamic multihop route planning for goods, as well as the effectiveness of combined load transportation (goods + passengers) as opposed to independent load transportation.

Our proposed method, \textit{Combined Load (DARM + DPRS + DMH)}, incorporates both multihop route planning and combined load delivery with pricing strategies. As compared to \textit{Combined Load (DARM + DPRS)} baseline, we hypothesize that the multihop route planning would be a more effective approach. Given that the core intuition of multihop route planning is to break trips into smaller segments and delivery only to hop locations that are convenient for the vehicle, we expect to see improvements in the number of requests served, the number of fleet vehicles deployed, profits, cruising time, travel distance, and occupancy rate. 

Likewise, we have included baselines that do not consider combined load transportation to observe the effectiveness of the combined load transportation. In these scenarios, we formulate our hypothesis that a combined load transportation scenario improves fleet utilization due to the more efficient packing of the vehicle. As opposed to independent load scenarios, in the combined load scenario, we utilize the vehicle trunk capacity. Additionally, the flexibility a vehicle has in transporting both passengers and goods should provide improved profits in the combined scenario.

The combination of these two strategies was expected to yield the most improved fleet utilization and profits.  {In addition, we verify the effectiveness of the DPRS framework, where drivers negotiate the price with passengers, by comparing against a baseline that does not adopt DPRS in Fig. \ref{fig:metrics_DPRS}. Note that, for goods, only the initial pricing is used and no negotiation takes place. Further, we verify the multi-hop routing for goods (DMH) by comparing against a baseline that doesn't adopt multi-hops (in Fig. \ref{fig:metrics_hops}), instead, this baseline only drops the goods at their final destination. Further, we see in Appendix \ref{apd_eval} that the dispatch decision takes $< 0.2$ seconds. Impact of reject radius and drop-off radius is also investigated in Appendix \ref{apd_eval}.}

\vspace{-0.1in}
\subsection{Results Discussion}
From our simulations, we observe that the hypothesis for each baseline comparison has been supported for the most part by our experimental results. 

Fig. \ref{fig:metrics_DPRS} shows that the distributed pricing framework (DPRS) improves the number of accepted passengers' requests by $10 - 30\%$, decreases the average travel distance of the fleet by about $10-25\%$, and saves up to $800$ vehicles from the fleet to serve other requests (i.e., the occupancy rate of the fleet) when compared to a non-pricing baseline.  This proves that DPRS framework is able to reach a common ground between passengers and drivers to accommodate both their preferences. In addition,  Fig. \ref{fig:metrics_hops} shows that the dynamic multi-hop approach (DMH) increases the accept rate of goods' requests by $15-20 \%$, decreases the travel distance of the fleet by $10-20\%$, and the occupancy rate by about $500$ vehicles on average, when compared to a non multi-hop baseline. This proves the efficiency of the DMH approach where it serves more requests with a lower number of vehicles and without increasing their average travel distance.

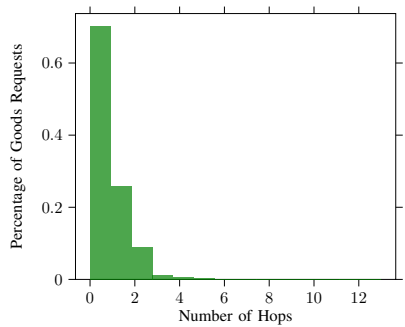
\begin{figure}[ht]
	\centering
	\resizebox{0.3\textwidth}{!}{  
\begin{tikzpicture}

\begin{axis}[
tick align=outside,
tick pos=both,
x grid style={white!69.0196078431373!black},
xlabel={Number of Hops},
xmin=-0.65, xmax=13.65,
xtick style={color=black},
y grid style={white!69.0196078431373!black},
ylabel={Percentage of Goods Requests},
ymin=0, ymax=0.737129666959457,
ytick style={color=black}
]
\draw[draw=none,fill=green!50!black,fill opacity=0.7] (axis cs:0,0) rectangle (axis cs:0.928571428571429,0.702028254247102);
\draw[draw=none,fill=green!50!black,fill opacity=0.7] (axis cs:0.928571428571428,0) rectangle (axis cs:1.85714285714286,0.258033908318665);
\draw[draw=none,fill=green!50!black,fill opacity=0.7] (axis cs:1.85714285714286,0) rectangle (axis cs:2.78571428571429,0.0889510759900076);
\draw[draw=none,fill=green!50!black,fill opacity=0.7] (axis cs:2.78571428571429,0) rectangle (axis cs:3.71428571428571,0.0120779648091334);
\draw[draw=none,fill=green!50!black,fill opacity=0.7] (axis cs:3.71428571428571,0) rectangle (axis cs:4.64285714285714,0.00624836534004335);
\draw[draw=none,fill=green!50!black,fill opacity=0.7] (axis cs:4.64285714285714,0) rectangle (axis cs:5.57142857142857,0.00359302694550518);
\draw[draw=none,fill=green!50!black,fill opacity=0.7] (axis cs:5.57142857142857,0) rectangle (axis cs:6.5,0.00219950661786961);
\draw[draw=none,fill=green!50!black,fill opacity=0.7] (axis cs:6.5,0) rectangle (axis cs:7.42857142857143,0.00137459305663203);
\draw[draw=none,fill=green!50!black,fill opacity=0.7] (axis cs:7.42857142857143,0) rectangle (axis cs:8.35714285714286,0.000893524918625409);
\draw[draw=none,fill=green!50!black,fill opacity=0.7] (axis cs:8.35714285714286,0) rectangle (axis cs:9.28571428571429,0.000545736313935378);
\draw[draw=none,fill=green!50!black,fill opacity=0.7] (axis cs:9.28571428571428,0) rectangle (axis cs:10.2142857142857,0.000392740873323437);
\draw[draw=none,fill=green!50!black,fill opacity=0.7] (axis cs:10.2142857142857,0) rectangle (axis cs:11.1428571428571,0.000271290884384061);
\draw[draw=none,fill=green!50!black,fill opacity=0.7] (axis cs:11.1428571428571,0) rectangle (axis cs:12.0714285714286,0.000182174983409064);
\draw[draw=none,fill=green!50!black,fill opacity=0.7] (axis cs:12.0714285714286,0) rectangle (axis cs:13,0.000130913624441146);
\end{axis}

\end{tikzpicture}}
	\caption{Histogram of number of hops for all goods deliveries.}\label{fig:n_hops}
\end{figure}

On the proposed method, we observe the number of hops taken by goods on Figure \ref{fig:n_hops} where 25\% of requests had one stop and 8\% had two stops. Looking at Figure \ref{fig:accept_sep} and Fig. \ref{fig:reject_sep} (given in Appendix \ref{ablation}), we notice that our proposed method shows the highest number of requests accepted over the course of the simulation (serving $\approx 97\%$ of the total goods' requests and up to $99\%$ of the total passengers requests). \textit{Combined Load (DARM + DPRS)} comes second, while the two baselines which do not consider a combined load come slightly behind. What is most noteworthy, however, is that the proposed method is able to deliver more requests while having the fewest number of occupied vehicles.We notice approximately 50\% improvement relative to \textit{Independent Load (DARM + DPRS)}, approximately 30\% improvement from \textit{Independent Load (DARM + DPRS + DMH)}, and about 10\% improvement to \textit{Combined Load (DARM + DPRS)}.  {Further, Fig. \ref{fig:wait_cap} (in Appendix \ref{ablation}) shows that the proposed method improves the utilization of both the seating and trunk capacities, with about $22 \%$ occupied trunk capacity and $40 \%$ occupied seating capacity on average, which leaves enough room for serving future requests, owing in part to the use of multihop route planning for goods.  In addition, it shows that the average waiting time for goods' doesn't exceed $20$ seconds; whereas for passengers, it is between $10 - 15$ seconds.  Further analysis of the baselines shown in these figures will be discussed in Appendix \ref{ablation} as we investigate the effectiveness of each component of our framework.}


In summary, our proposed method combines two strategies that have demonstrated a more profitable and efficient fleet usage: multihop route planning, and combined load transportation. With the addition of insertion based route planning, and demand aware pricing strategies proposed in \cite{haliem2020distributed}, we demonstrate effective strategies for the efficient management of delivery fleet vehicles for both goods and passengers. 

\section{Conclusions and Future Work}\label{Conclusions}

In this paper, we proposed a framework to enable distributed decision making for fleet vehicles with passenger and goods delivery tasks with the objective of maximizing operational profitability and customer convenience. The system includes a novel route-planning and matching algorithm, pricing strategies, and a Deep Reinforcement Learning based dispatch algorithm.   Multi-hop and insertion-cost considerations in our route planning enable vehicles to generate ideal routes on-the-fly. Our pricing strategy which is part of the planning mechanism involves both customers as well as drivers in the decision-making processes. The agents are informed by utility functions that aim to achieve a maximum monetary profit for both drivers and customers. We integrate these algorithms with a distributed DQN approach which dispatches idle vehicles with the objective of anticipating profits from potential customers across the map thus maximizing profit while minimizing idle driving and fuel expenses. Evaluation in a realistic discrete-event simulator demonstrates promising results for the inclusion of multi-hop considerations in route planning. Most notably, we observed a significant reduction in idle driving time which saves on fuel expenses and improved fleet utilization by approximately 15\%. The 20\% improvement in profits also points to great potential for adoption in a profit-driven society. It is also worth noting that our approach can easily be scaled up to a large number of fleet vehicles and readily deployable locally within each vehicle to enable distributed inference.

In regards to future research directions, we propose developing systems that can incorporate deadline-based constraints to enable transportation of highly urgent goods such as medicines or perishable items. Additionally, it would be interesting to see the improvement of the proposed approach in less densely populated locations than New York City. Extensions to account for non-stationarity \cite{adapool,qian2017time} and use of intelligent option policies \cite{drop} in this framework are important research directions. 


\bibliographystyle{ieeetr}
\bibliography{references}

\newpage
\clearpage

\appendices
\section{Dataset}

The delivery workload in this simulator consists of passenger pick-up requests as well as package delivery requests for various goods and services. 
To emulate a realistic workload of passenger pick-ups in the urban environment, we used the New York City taxi trip data set from \cite{taxi2018limousine}. We extracted trips within the major burrows of the metropolitan area from May and June for our simulation.

\begin{figure}
	\includegraphics[width=7.7cm]{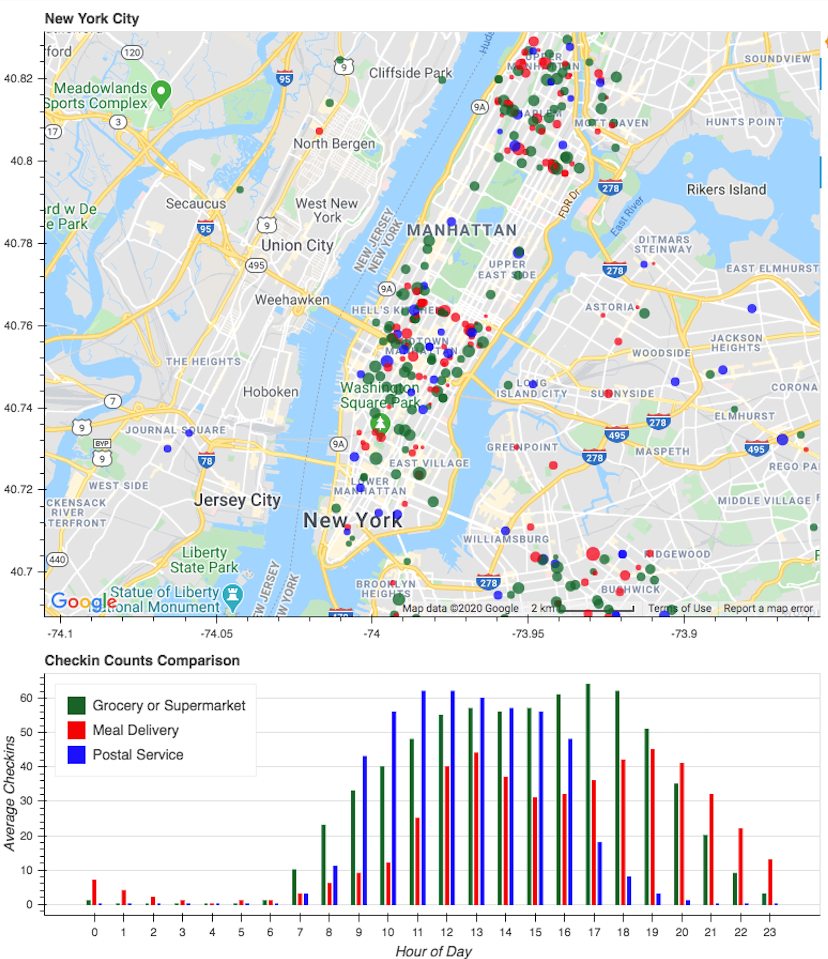}
	\caption{The map and graph show different types of goods delivery requests.}
	\label{fig:Checkins}
\end{figure}
In simulating package delivery requests, customer check-in traffic was extracted from Google Maps for postal service, meal delivery, and supermarket locations. The 100 most active locations were considered from each of the respective service types. Average check-in traffic was extracted for each day of the week for each location. This was used to generate a synthetic workload for a total of two months representative of May and June. Figure \ref{fig:Checkins} shows the distribution of check-ins across the city over a synthetic workload, which was generated from customer check-in from Google Maps analytics data. At each service location, the request rates were generated by Poisson distribution given in Equation (\ref{Poisson}) across request rate $x$,  where $\lambda$ represents the observed check-in rate from Google Maps. Consequently, for each service location, package drop-off locations were generated randomly considering delivery radius limit of 5 miles in accordance with the current standard for major crowd-sourced delivery services such as DoorDash and GrubHub. All pick-up and drop-off locations are constrained within the New York City borough boundaries. The resulting data set consisted of goods delivery orders over one month. 

\begin{equation}
p(x; \lambda) = \frac{e^{-\lambda} \lambda^x}{x!}\\
; x \in \mathbb{Z}
\label{Poisson}
\end{equation}
We let the vehicle carrying capacities be $C_p = 4$ passengers and $C_k = 5$ packages, unless stated otherwise. A request is deemed rejected if there are no vehicles available within a radius of $5km^2$. When an adequate number of vehicles are not present to meet the demand of pick-up requests, a higher reject rate can be observed.

\begin{figure}[htbp]
	\includegraphics[width=7.7cm]{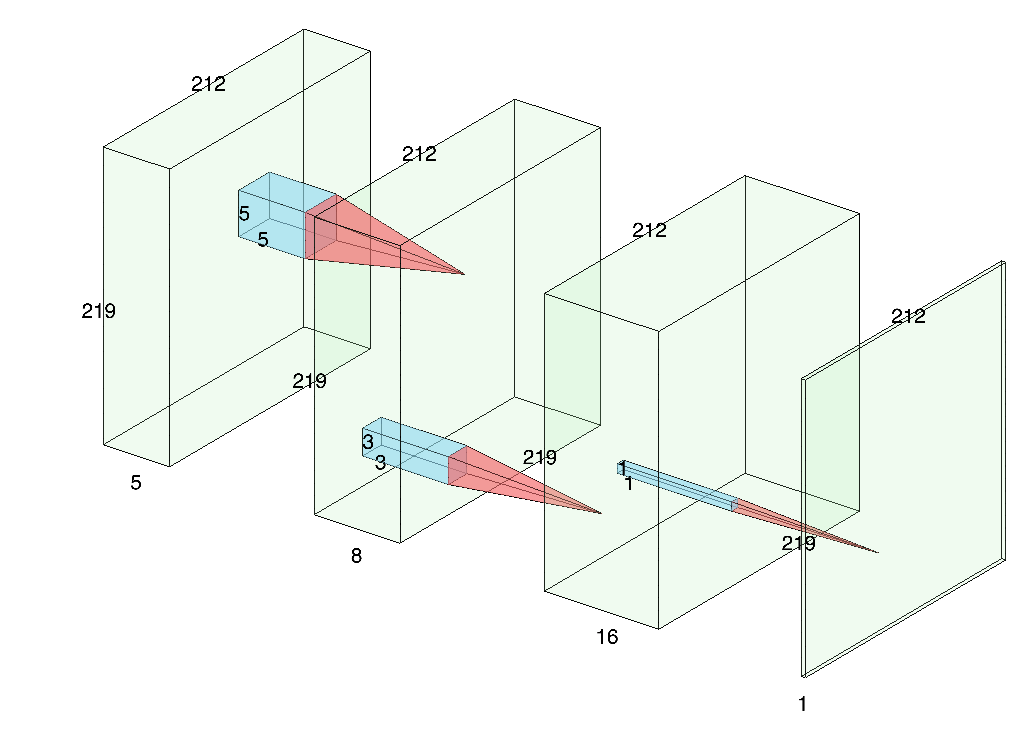}
	\caption{This diagram shows the Conv-Net architecture of the Demand Prediction Model.}
	\label{fig:Demand}
\end{figure}

\begin{figure*}[htbp]
	\begin{subfigure}{.48\textwidth}
		\centering
		\includegraphics[width=7.7cm]{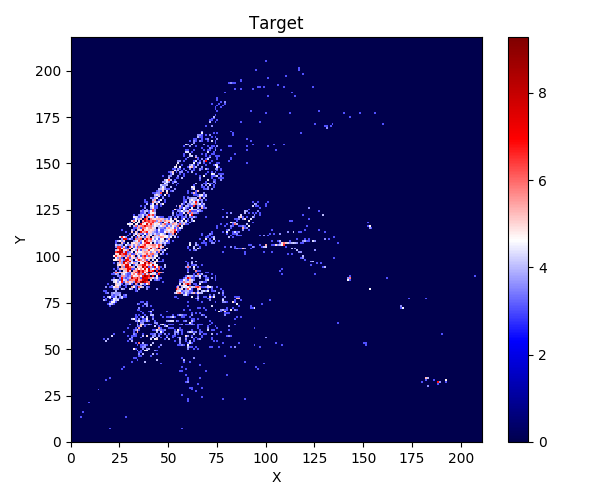}
		\caption{This shows a heatmap of actual demand from the dataset which is used as a target to train the demand prediction model.} \label{fig:1a}
	\end{subfigure}
	\begin{subfigure}{.48\textwidth}
		\centering
		\includegraphics[width=7.7cm]{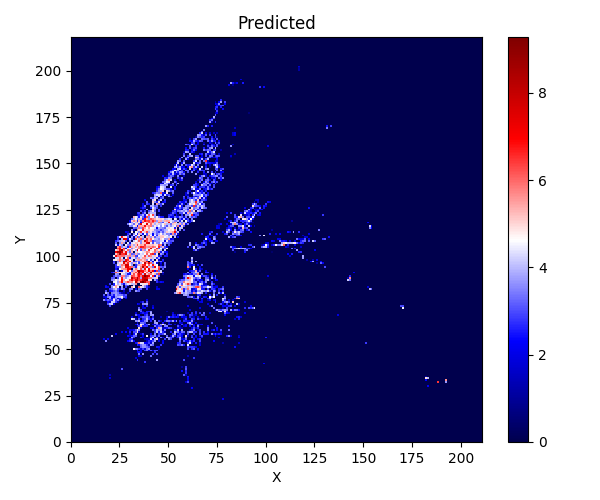}
		\caption{This shows a heatmap obtained from the demand prediction model which predicted pickup demand over the map.} \label{fig:1b}
	\end{subfigure}
	\caption{The above heatmaps visualize target and predicted demand areas. Warmer colors indicate a higher demand as shown by the color scale.} \label{fig:Demand_Heatmap}
\end{figure*}

\section{ETA and Demand Prediction}\label{eta_pred}

The simulator uses an Estimated Time of Arrival (ETA) model to predict the estimated trip times. This model is built using the New York City taxi data set. In the ETA model, we want to predict the expected travel time between two zones (two pairs of latitudes and longitudes). We split our data into 70\% train and 30\% test. We use day of week, latitude, longitude and time of days as the explanatory variables and use random forest to predict the ETA. The final ETA model yielded a root mean squared error (RMSE) of 3.4 on the test data.

The demand prediction model is a critical element to the simulator in building the state space vector that allows DQN agents to proactively dispatch towards areas where there is a high demand. This model is built using the Conv-Net architecture shown in Figure \ref{fig:Demand}. The network outputs a  $212 \times 219$  heat  map image in which each pixel stands for the predicted number of pick-up requests for each location on the map for the following 30 minutes of simulation. The network is fed with input images which represent the actual pick-ups of all service types over the last 6 time-steps. The actual pick-up counts over the map is combined with the sine and cosine of the day of week and hour of day to capture the daily and weekly periodicity of the demand. After training using a 80\% train and 20\% test split, RMSE values for training and testing were 0.945 and 1.217 respectively. Figure \ref{fig:Demand_Heatmap} shows the demand heat map of a target sample and a predicted sample of this demand prediction model. The predicted demand in this figure is a sample $212 \times 219$ output of the Conv-Net Architecture from Figure \ref{fig:Demand}.

\section{DARM Distributed Optimization for Route Planning}\label{apd:darm}

This demand-aware route planning problem is a variation of the basic route planning problem for shareable mobility services \cite{tao2007dynamic,tong2018unified}. Similar to the basic route planning problem \cite{tao2007dynamic,tong2018unified}, demand aware route planning is also an NP-hard problem. As proved in \cite{tong2018unified} there is no optimal method to maximize the total revenue for the basic route planning problem (which is reducible to demand aware route planning) using neither deterministic nor randomized algorithms. The same applies to DARM as a result. Even though there is no polynomial time algorithm to optimally solve the demand-aware route planning problem, we utilize an insertion based approach as it has been proven to be a effective in the shared transportation space \cite{tong2018unified, haliem2020distributed}.  

DARM performs a local optimization on each route by inserting new vertices into the vehicle's route (which is represented as a Directed Graph). We define the insertion operation as: given a vehicle $V_{j}$ with the current route $S_{V_{j}}$,  and a new request $r_{i}$ with two vertices (i.e., origin $o_{i}$ and destination $d_{i}$), the insertion operation aims to find a new feasible route $S^{\prime}_{V_{j}}$ by inserting $o_{i}$ and $d_{i}$ into $S_{V_{j}}$ with the minimum increased cost, that is the minimum extra travel distance, while maintaining the order of vertices in $S_{V_{j}}$ unchanged in $S^{\prime}_{V_{j}}$. For each $r_{i}$, the basic insertion algorithm checks every possible position to insert the origin and destination locations and return the new route such that the incremental cost is minimized. In presenting the cost function, we first define our distance metric, where given a graph $G$ we use our routing engine to pre-calculate all possible routes over our simulated city. We then derive the distances of the paths from location $a$ to location $b$ to define our graph weights. Consequently, we obtain a weighted graph $G$ with realistic distance measures serving as its weights. The weight notation to paths is presented as:
$w(a_{1}, a_{2}, ..., a_{n}) = \sum_{i=1}^{n-1} w(a_{i}, a_{i+1})$.

\begin{algorithm}
	\caption{Insertion-based Route Planning}\label{alg:route_planning}
	\begin{algorithmic}[1]
		\State \textbf{Inputs}: Vehicle $V_j$, its current route $S_{V_j}$, and request  $r_i = (o_i,d_i)$ and weighted graph G
		\State \textbf{Outputs}: Route $S'_{V_j}$ after insertion and minimization of $cost(V_j, S'_{V_j})$
		\If {$S_{V_j}$ is empty}
			\State $S_{V_j} \gets [loc(V_j),o_i,d_i].$,
			\State $cost(V_j,S'_{V_j}) = \omega(S'_{V_j})$
			\State \textbf{Return} $S'_{V_j}$, $cost(V_j, S'_{V_j})$
		\EndIf
		\State \textbf{Initialize} $S''_{V_j} = S_{V_j}$, $Pos[o_i] = NULL$, $cost_{min} = +\infty$
		\For {each x in 1 to $|S_{V_j}|$}
			\State $S^x_{V_j}$ := \textbf{Insert} $o_i$ at $x$-th in $S_{V_j}$
			\State \textbf{Calculate} $cost(V_j, S^x_{V_j}) = \omega(S^x_{V_j})$
			\If {$cost(V_j, S^x_{V_j}) < cost_{min}$}
				\State $cost_min \gets cost((V_j, S^x_{V_j})$
				\State $Pos[o_i] \gets x$ , $S''_{V_j} \gets S^x_{V_j}$
			\EndIf
		\EndFor
		\State $S'_{V_j} = S''_{V_j}, cost_{min} = +\infty$
		\For {each y in $Pos[o_i] + 1$ to $|S'_{V_j}|$}
			\State $S^x_{V_j}$ := \textbf{Insert} $o_i$ at $y$-th in $S'_{V_j}$
			\State \textbf{Calculate} $cost(V_j, S^y_{V_j}) = \omega(S^y_{V_j})$
				\If {$cost(V_j, S^y_{V_j}) < cost_{min}$}
				\State $cost_min \gets cost((V_j, S^y_{V_j})$
				\State $S'_{V_j} \gets S^y_{V_j}$ , $cost(V_j, S'_{V_j}) \gets cost_{min}$
			\EndIf
		\EndFor
		\State \textbf{Return} $S'_{V_j}, cost(_{V_j, S'_{V_j}})$
	\end{algorithmic}
\end{algorithm}

We define the cost associated with each new potential route $S^{\prime}_{V_{j}} = [r_{i}, r_{i+1}, ..., r_{k}]$ to be the $\text{cost}(V_{j}, S^{\prime}_{V_{j}}) = w(r_{i}, r_{i+1}, ... r_{k})$ resulting from this specific ordering of vertices (origin and destination locations of the $k$ requests assigned to vehicle $V_{j}$). We derive the cost of the original route to calculate the increased costs for every new route $S^{\prime}_{V_{j}}[r_{i}]$. To illustrate this, we assume $A_{j}$ for vehicle $V_{j}$ has only two requests $r_{x}$ and $r_{y}$, its location is $loc(V_{j})$ and its current route has $r_{x}$ already inserted as: $[loc(V_{j}), o_{x}, d_{x}]$. Then, $V_{j}$ picks $S^{\prime}_{V_{j}}$ of inserting $r_{y}$ into its current route, by the following cost function:
$
\text{cost}(V_{j}, S^{\prime}_{V_{j}}[r_{y}] ) = \text{min} [ w(loc(V_{j}), o_{x}, o_{y}, d_{x}, d_{y}), \\
w(loc(V_{j}), o_{y}, o_{x}, d_{x}, d_{y}),
w(loc(V_{j}), o_{y}, o_{x}, d_{y}, d_{x}),\\
w(loc(V_{j}), o_{x}, o_{y}, d_{y}, d_{x}), 
w(loc(V_{j}), o_{x}, d_{x}, o_{y}, d_{y}), \\
w(loc(V_{j}), o_{y}, d_{y}, o_{x}, d_{x}) ].
$
Note that the last two optional routes complete one request before serving the other, hence they do not fit into the \textit{ride-sharing} category. However, we still take them into consideration we optimize for the fleet's overall profits and total travel distance. Also, note that these two routes will still serve both requests and thus would not affect the overall acceptance rate of our algorithm. Note that if this was the first allocation made to this vehicle, then the first request will be just added to its currently empty route. Otherwise, the first request will be dealt with (like all other requests in the list) by following the insertion operation described above. 
As each vehicle minimizes its own travel cost, this optimization works in a distributed fashion following Algorithm \ref{alg:route_planning}. This distributed procedure is performed for each request $r_i$ as follows:
\begin{itemize}[leftmargin=*]
	\item Each vehicle $V_{j}$ receives its initial request assignments $A_{j}$ along with an price $P_{init}$ (as explained in Section \ref{Initial_Pricing}) associated with each request $r_{i}$ in that list. The assigned requests are sorted in ascending order of proximity to vehicle $V_{j}$.
	\item Each vehicle $V_{j}$ then considers the requests in this ordered list by inserting one at a time into its current route. For each request $r_{i}$, the vehicle arrives at the minimum cost  $\text{cost}(V_{j}, S^{\prime}_{V_{j}}[r_{i}] )$ associated with inserting it to its current route (as described above). 
	\item If $r_i$ is a goods request, the vehicle sticks to the initial price as it performs its insertion cost procedure only to identify the most convenient path in dropping off the package.
	\item If $r_i$ is a passenger, given the initial price associated with this request $P_{init}$ and the the new route $S^{\prime}_{V_{j}}[r_{y}]$ which may involve detours to serve this request, the vehicle will re-calculate the pricing taking into consideration any extra distance using Equation (\ref{eqn:update_price}) in Section \ref{Vehicle_Proposed_Pricing}. 
	\item For passengers, the vehicles will then modify the pricing based on the Q-values of the driver's dispatch-to location, having gained that insight using the DQN dispatch policy about which destinations can yield him/her a higher profit. The drivers weigh in on their utility function and propose a new pricing to the customer using Equation \eqref{eqn:update_price}. This procedure is explained in Section \ref{Vehicle_Proposed_Pricing}.
	\item Finally, the customer(s) can accept or reject based on his/her utility function as explained in Section \ref{Passenger_Decision_Function} If a customer accepts, the vehicle updates its route $S_{V_{j}}$ to be $S^{\prime}_{V_{j}}[r_{y}]$, otherwise $S_{V_{j}}$ remains unchanged. The vehicle then proceeds to the next customer and repeats the process. This rejected request will be fed back into the system to be considered in the matching process initiated  in the next timestep for other/same vehicles.
\end{itemize}
\section{Additional Evaluations}\label{apd_eval}
\begin{figure}
	\centering
	\includegraphics[width=0.4\textwidth]{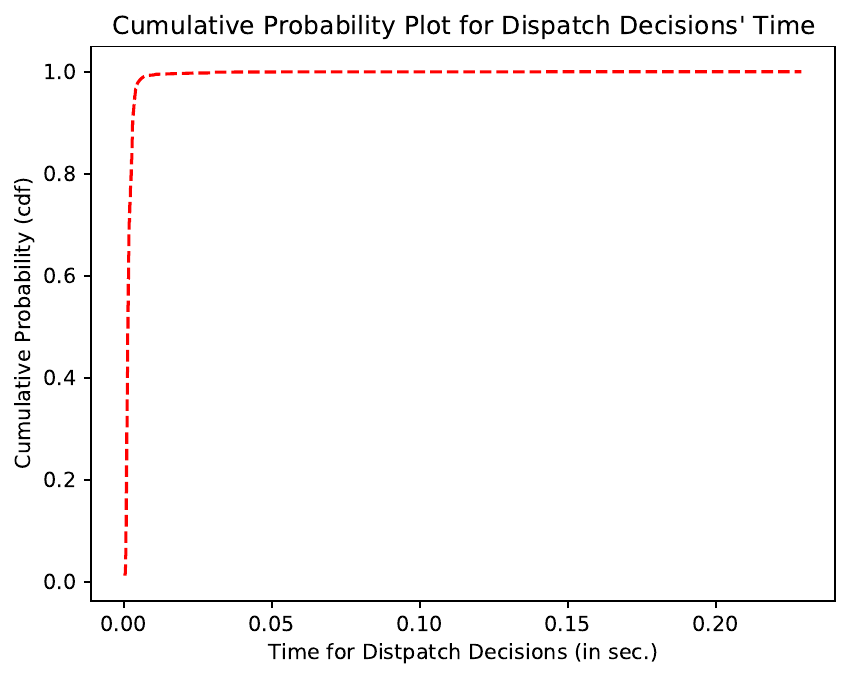}
	\vspace{-0.1in}
	\caption{CDF for Dispatch Decisions’ Time}\label{dis_time}
\end{figure}
\begin{figure*}
\centering
\begin{minipage}[t]{0.75\textwidth}
	\resizebox{\textwidth}{!}{  
		\includegraphics[trim = 70 0 70 0]{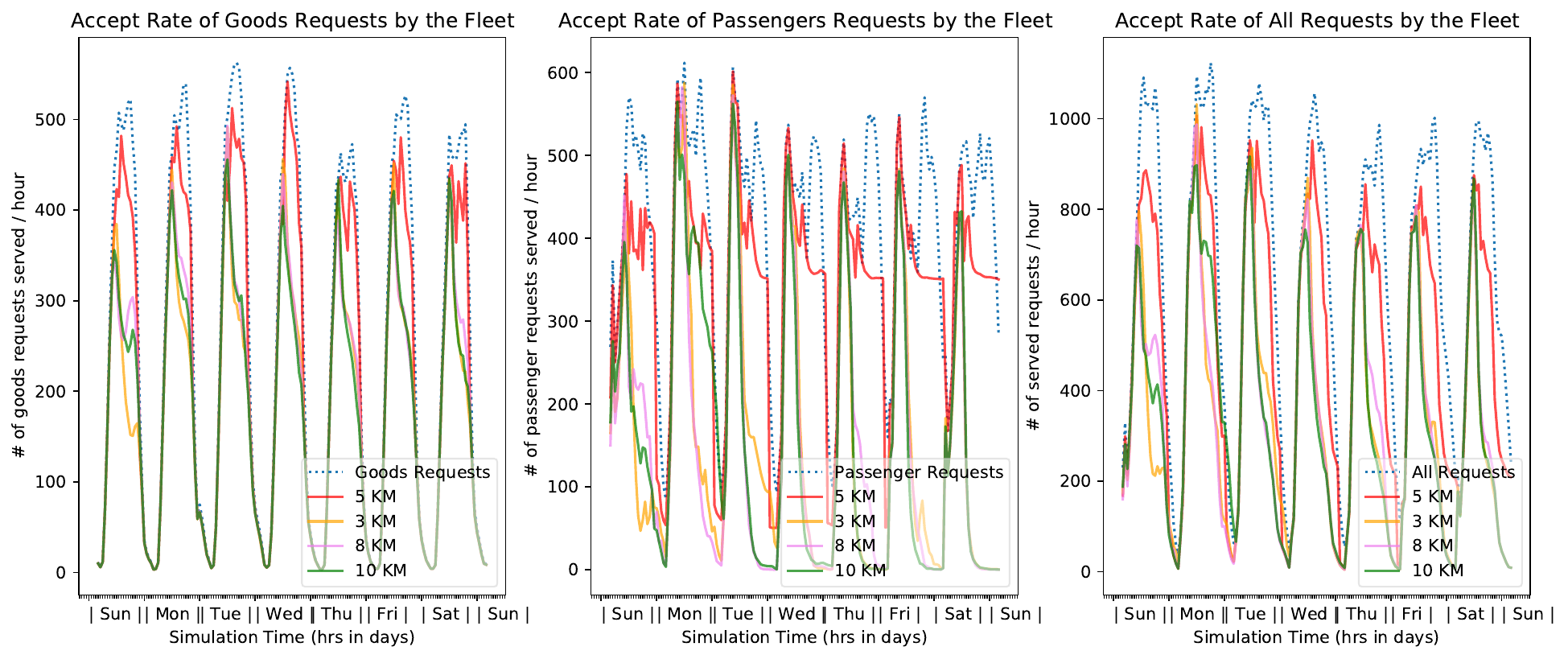}}
	\caption{Sensitivity Analysis for the Reject Radius Parameter.}\label{fig:m_radius}
\end{minipage}
\hspace{.01in}
\begin{minipage}[t]{0.75\textwidth}
	\resizebox{\textwidth}{!}{  
		\includegraphics[trim = 70 0 70 0]{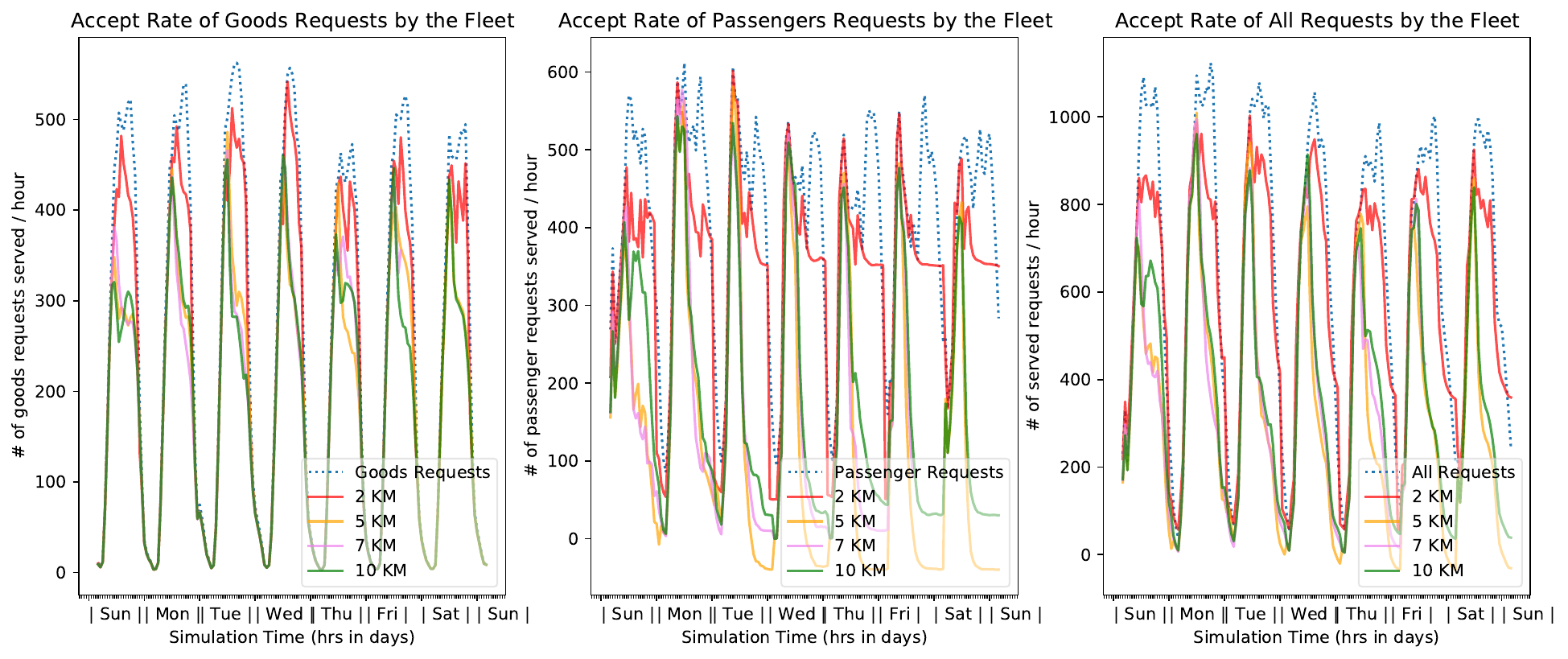}}
	\caption{Sensitivity Analysis for the Dropoff Radius Parameter.}\label{fig:d_radius}
\end{minipage}
\end{figure*}

\subsection{{Computational Analysis}}
{To provide more insight regarding the complexity of our framework, we investigate the time taken to make dispatch decisions. We show in Fig. \ref{dis_time}, the cumulative distribution function (cdf) plot for the time taken for dispatch decisions for each individual vehicle. We can observe that with probability 1.0, it will take the vehicle $< 0.2$ seconds to make a dispatch decision of which location on the map to head to next in order to maximize its own reward. }

\subsection{{Sensitivity Analysis}}
{In this section, we investigate the effect of the reject radius (requests are automatically rejected if no vehicle is available within this radius to serve the request) and the drop-off radius (Goods don't go through another hop if the distance to reach their final destination is within this radius) on the overall accept rate of the ridesharing system.  In Fig. \ref{fig:m_radius}, we vary the reject radius between $[3, 10]$ to observe how it affects the accept rates for passengers and goods, and thus for the overall system. We can conclude that a radius of $5 km$ reaches a balance between passengers' and goods' requests and thus, achieves the highest overall accept rate for our system. Likewise,  Fig. \ref{fig:d_radius} studies the effect of varying the dropoff radius between $[2, 10]$ on the overall acceptance rate of our system. We note that, there is a trade-off here as when a vehicle drops-off goods at a hop, this space in its trunk becomes available to serve another goods' request and thus it affects the number of served requests by the fleet.  Vehicles also aim to optimize their routing in order not to lose any passengers' requests en route due to large wait times. We can conclude that, with dropoff radius of $2 km$, the fleet is able to maximize the number of requests served for both goods and passengers.}

\section{{Ablation Study}} \label{ablation}
\begin{figure*}
	\centering
	\begin{minipage}[t]{0.9\textwidth}
	\resizebox{\textwidth}{!}{  
	\includegraphics[trim = 120 10 120 50]{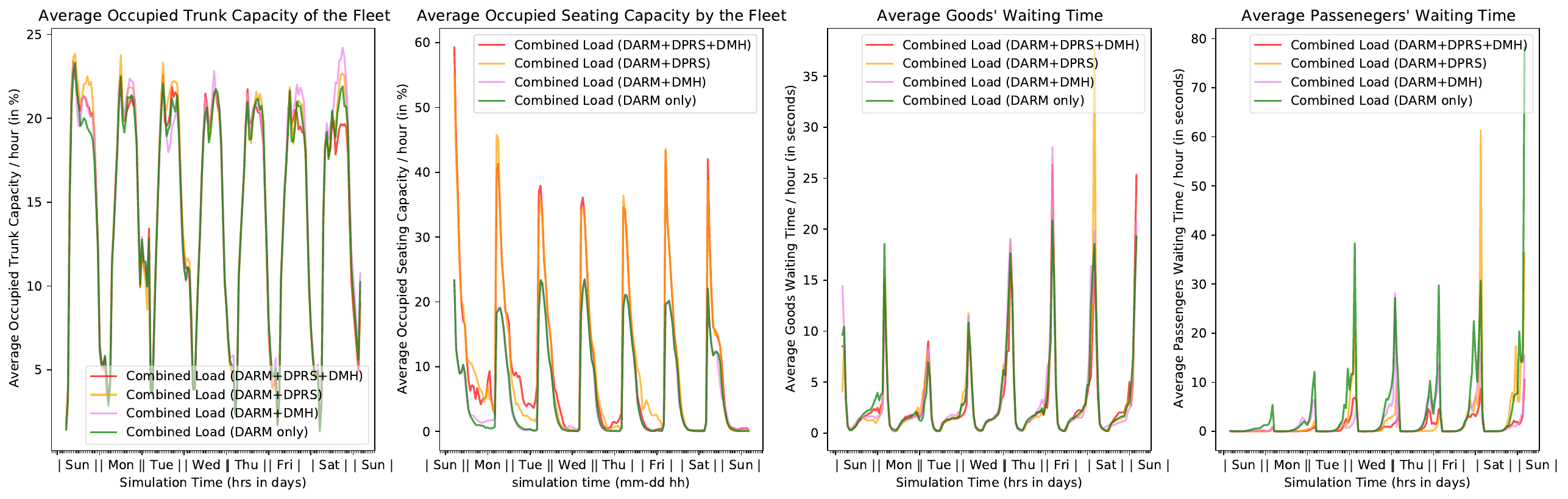}}
	\vspace{-0.1in}
	\caption{Occupancy and Wait Time for Passengers/Goods separately}\label{fig:wait_cap}
	\end{minipage}
	\begin{minipage}[t]{0.8\textwidth}
	\resizebox{\textwidth}{!}{  
	\includegraphics[scale = 0.4,trim = 70 10 70 0]{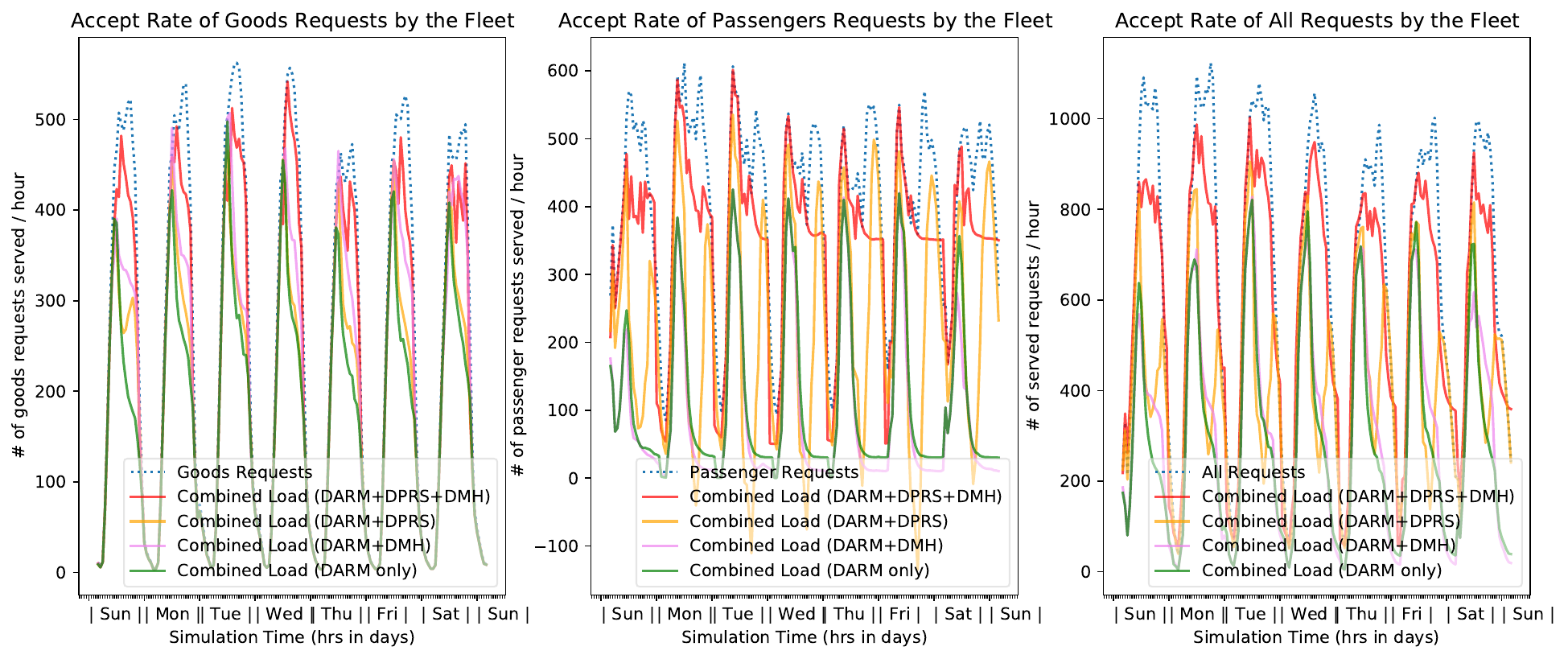}}
	\vspace{-0.1in}
	\caption{Accepted Requests for Passengers/Goods separately}\label{fig:accept_sep}
	\end{minipage}
	\begin{minipage}[t]{0.8\textwidth}
	\resizebox{\textwidth}{!}{  
	\includegraphics[scale = 0.4, trim = 70 10 70 0]{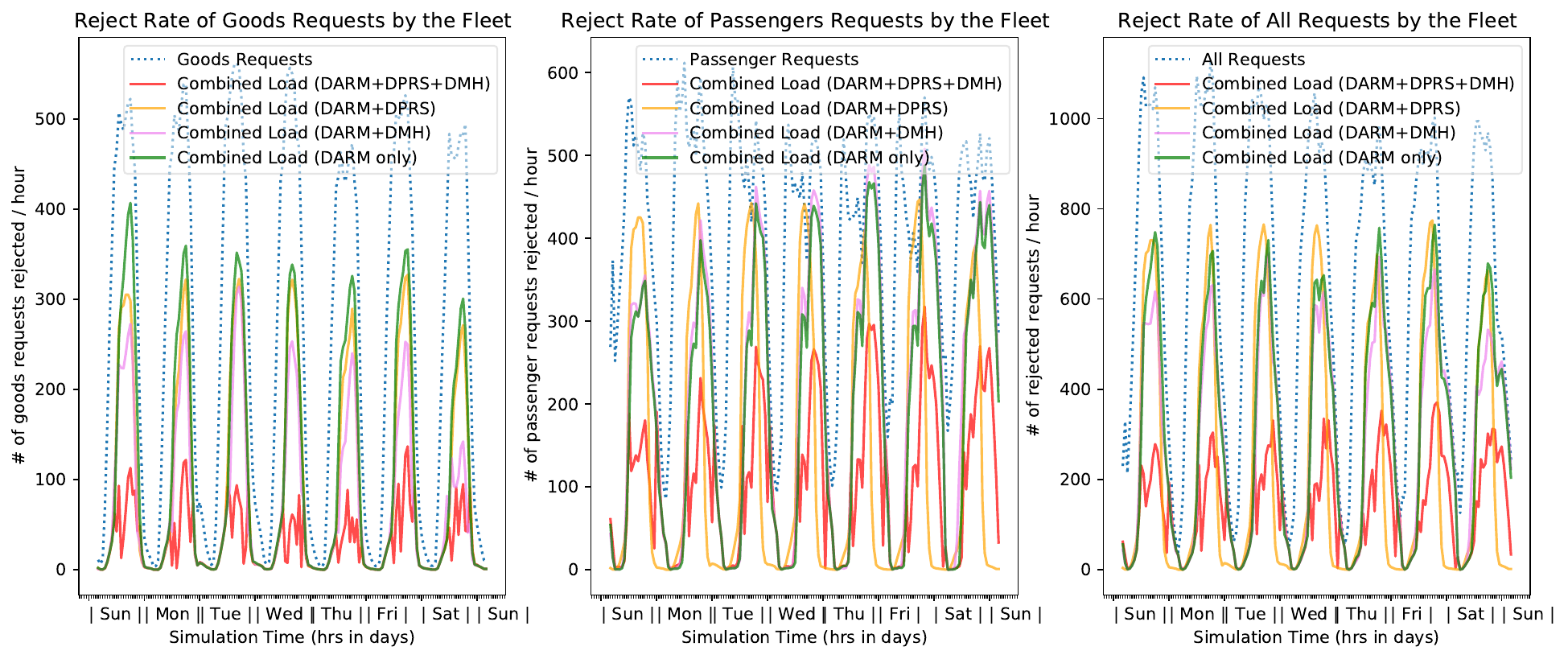}}
	\vspace{-0.1in}
	\caption{Rejected Requests for Passengers/Goods separately}\label{fig:reject_sep}
	\end{minipage}
\end{figure*}

{In this section, we investigate the effectiveness of each component of our framework where we show their impact on different performance metrics as shown in Figures \ref{fig:wait_cap}, \ref{fig:accept_sep}, and \ref{fig:reject_sep}.  For a combined load of passengers and goods, we compare a baseline with the insertion-based matching and route-planning (DARM) only, a baseline with DARM in addition to our dynamic multi-hop routing (DARM+DMH), a baseline with DARM in addition to the distributed pricing strategy (DARM + DPRS) against our proposed approach with all these components (DARM+DPRS+DMH). This study helps identify the impact of each component separately.  We can observe that even though the (DARM only) baseline achieves good utilization of both the trunk and seating capacities (as shown in Fig. \ref{fig:wait_cap}), it serves only $50 \%$ of requests (Fig. \ref{fig:accept_sep}) with the highest reject rate for both passengers and goods (Fig. \ref{fig:reject_sep}), and results in a significantly higher waiting times for both passengers and goods.  On the other hand, both the (DARM + DMH) and the (DARM + DPRS) baselines achieve an improvement on all these metrics with the (DARM + DMH) improving the goods' accept rate, wait time and trunk capacity utilization, ans the (DARM + DPRS) baseline improving the passengers' accept rate, wait time and the seating capacity utilization. Clearly, our proposed approach combining both approaches (DARM+DPRS+DMH) ranks highest in improving both the utilization of the seating and trunk capacities as well as achieving the minimal waiting time for both goods and passengers. Finally, our proposed method serves the most number of requests ($\approx 97 \%$ of the total goods' requests and up to $99 \%$ of the total passengers requests), while minimizing the number of occupied vehicles of the fleet.
}

\begin{figure*}
	\centering
	\begin{minipage}[t]{0.8\textwidth}
		\resizebox{\textwidth}{!}{  
			\includegraphics{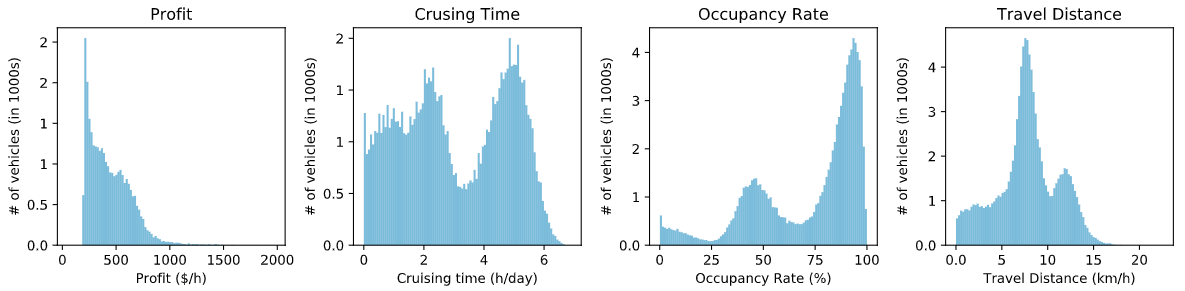}}
		\caption{\small Performance metrics for the Combined Load (DARM + DPRS + DMH) proposed method.}\label{fig:metrics_CMH}
	\end{minipage}
	\hspace{.01in}
	\begin{minipage}[t]{0.8\textwidth}
		\resizebox{\textwidth}{!}{  
			\includegraphics{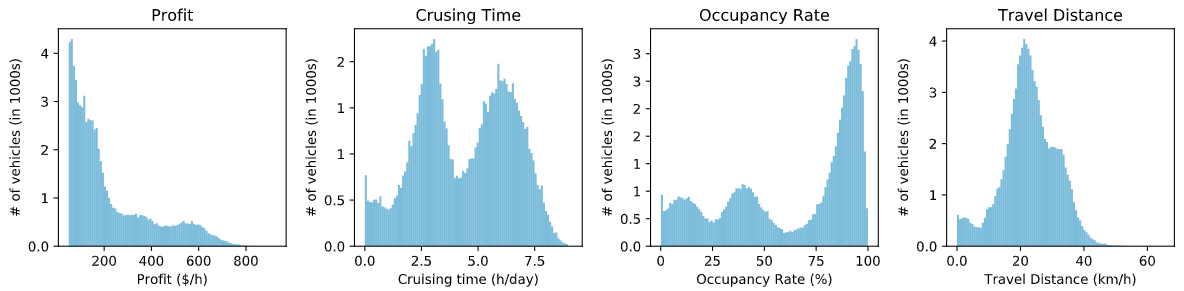}}
		\caption{\small Performance metrics for the Combined Load (DARM + DPRS) baseline.}\label{fig:metrics_CNH}
	\end{minipage}
	\hspace{.01in}
	\begin{minipage}[t]{0.8\textwidth}
		\resizebox{\textwidth}{!}{  
			\includegraphics{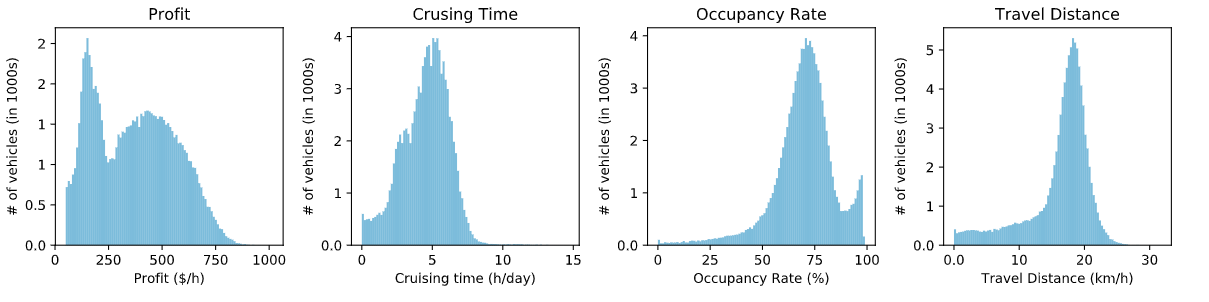}}
		\caption{\small Performance metrics for the Independent Load (DARM + DPRS + DMH) baseline. }\label{fig:metrics_IMH}
	\end{minipage}
	\hspace{.01in}
	\begin{minipage}[t]{0.8\textwidth}
		\resizebox{\textwidth}{!}{  
			\includegraphics{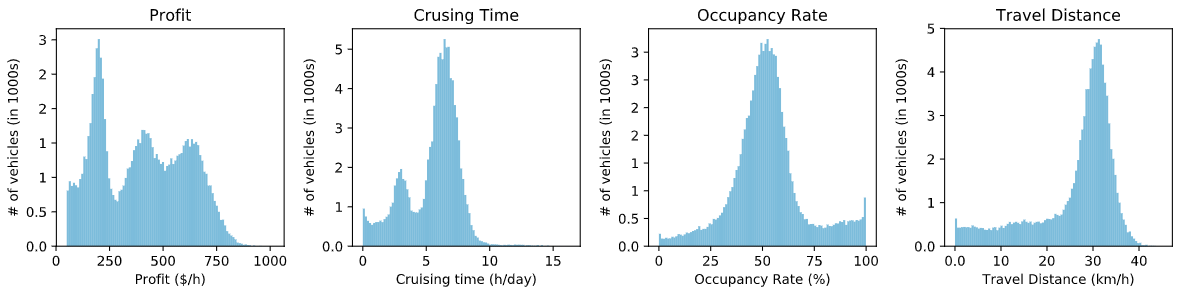}}
		\caption{\small Performance metrics for the Independent Load (DARM + DPRS) baseline.}\label{fig:metrics_INH}
	\end{minipage}
\end{figure*}

To evaluate the performance of each of the models further in terms of efficiency metrics stated earlier, we observe the plots shown in Figures \ref{fig:metrics_CMH}, \ref{fig:metrics_CNH}, \ref{fig:metrics_IMH}, and \ref{fig:metrics_INH}. In comparing methods that consider multihop routing (Figure \ref{fig:metrics_CMH},  Figure \ref{fig:metrics_IMH}) to their direct transit counterparts (Figure \ref{fig:metrics_CNH}, Figure \ref{fig:metrics_INH}), we first observe that multihop routing significantly reduces cruising time and travel time. As a result, we also see that the occupancy rate is improved for methods considering multihop routing. Our proposed method achieves the best for all these efficiency metrics. It is also to be noted that multihop routing improves the potential profit of vehicles. Comparing Figure \ref{fig:metrics_CMH} to Figure \ref{fig:metrics_CNH}, we see a potential profit gain for vehicles of approximately 20\%. With multihop routing, vehicles are able to deliver more requests to hop-zones which are conveniently located close to their current paths. This allows them to incur less cost and gain more profit for delivering requests. In evaluating the efficiency of combined load transportation, we compare Figure \ref{fig:metrics_CMH} and Figure \ref{fig:metrics_CNH} to Figure \ref{fig:metrics_IMH} and Figure \ref{fig:metrics_INH}, respectively. It can be seen from this comparison that the profits are significantly better with combined load transportation, in addition to a better occupancy rate.

\end{document}